\documentclass[letterpaper]{article} 
\usepackage{aaai25}  
\usepackage{times}  
\usepackage{helvet}  
\usepackage{courier}  
\usepackage[hyphens]{url}  
\usepackage{graphicx} 
\usepackage{natbib}  
\usepackage{caption} 
\usepackage{algorithm}
\usepackage{algorithmic}
\usepackage{subfig}

\usepackage{amsmath}
\usepackage{amsthm}
\usepackage{amssymb}
\usepackage{booktabs}
\usepackage{multirow}
\usepackage{newfloat}
\usepackage{listings}
\DeclareCaptionStyle{ruled}{labelfont=normalfont,labelsep=colon,strut=off} 
\floatstyle{ruled}
\newfloat{listing}{tb}{lst}{}
\floatname{listing}{Listing}
\title{Revisiting Graph Contrastive Learning on Anomaly Detection: A Structural Imbalance Perspective}

\author{
    Yiming Xu\textsuperscript{\rm 1,2},
    Zhen Peng\textsuperscript{\rm 1,2}\thanks{Corresponding author.},
    Bin Shi\textsuperscript{\rm 1,2},
    Xu Hua\textsuperscript{\rm 1,2},
    Bo Dong\textsuperscript{\rm 2,3},
    Song Wang\textsuperscript{\rm 4},
    Chen Chen\textsuperscript{\rm 4}
}
\affiliations{
    \textsuperscript{\rm 1}School of Computer Science and Technology, Xi'an Jiaotong University\\
    \textsuperscript{\rm 2}Shaanxi Provincial Key Laboratory of Big Data Knowledge Engineering, Xi’an Jiaotong University\\
    \textsuperscript{\rm 3}School of Distance Education, Xi’an Jiaotong University\\
    \textsuperscript{\rm 4}University of Virginia, Charlottesville, Virginia, USA\\

    \{xym0924, huaxu\}@stu.xjtu.edu.cn, zhenpeng27@outlook.com, \{shibin, dong.bo\}@xjtu.edu.cn, \{sw3wv, zrh6du\}@virginia.edu
}

\usepackage{bibentry}

\begin{document}

\maketitle

\begin{abstract}
The superiority of graph contrastive learning (GCL) has prompted its application to anomaly detection tasks for more powerful risk warning systems. Unfortunately, existing GCL-based models tend to excessively prioritize overall detection performance while neglecting robustness to structural imbalance, which can be problematic for many real-world networks following power-law degree distributions. Particularly, GCL-based methods may fail to capture tail anomalies (abnormal nodes with low degrees). This raises concerns about the security and robustness of current anomaly detection algorithms and therefore hinders their applicability in a variety of realistic high-risk scenarios. To the best of our knowledge, research on the robustness of graph anomaly detection to structural imbalance has received little scrutiny. To address the above issues, this paper presents a novel GCL-based framework named AD-GCL. It devises the neighbor pruning strategy to filter noisy edges for head nodes and facilitate the detection of genuine tail nodes by aligning from head nodes to forged tail nodes. Moreover, AD-GCL actively explores potential neighbors to enlarge the receptive field of tail nodes through anomaly-guided neighbor completion. We further introduce intra- and inter-view consistency loss of the original and augmentation graph for enhanced representation. The performance evaluation of the whole, head, and tail nodes on multiple datasets validates the comprehensive superiority of the proposed AD-GCL in detecting both head anomalies and tail anomalies. \footnote{The source code and datasets are available at: https://github.com/yimingxu24/AD-GCL}
\end{abstract}

\section{Introduction}

\begin{figure*}[t]
\centering
\subfloat[CoLA on Cora]{\label{fig:bias1}\includegraphics[width=0.25\linewidth]{
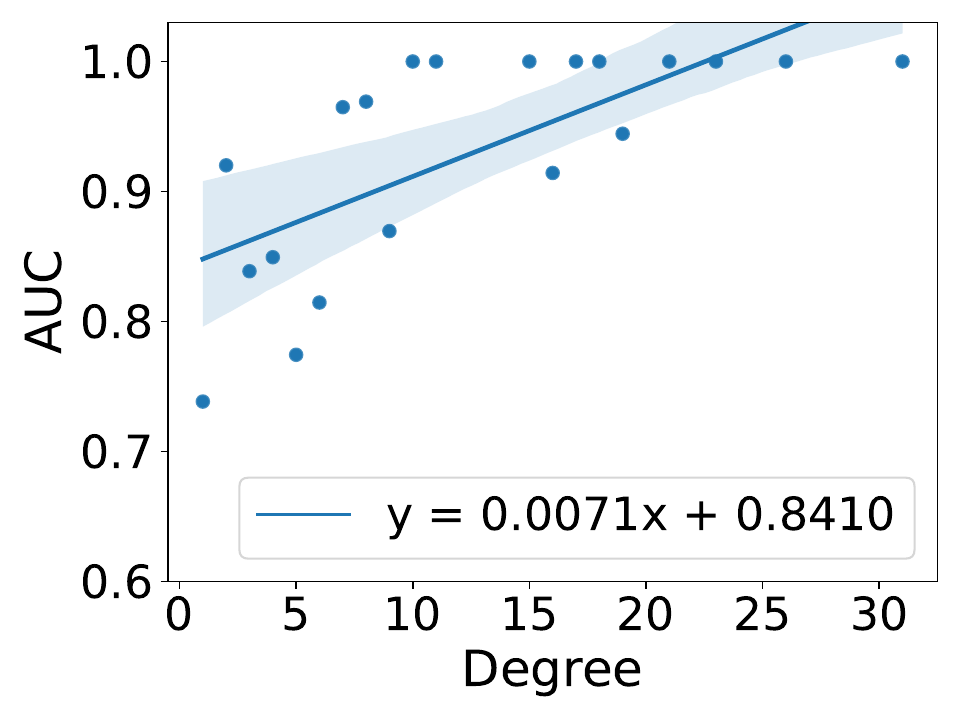}}
\subfloat[ANEMONE on Cora]{\label{fig:bias2}\includegraphics[width=0.25\linewidth]{
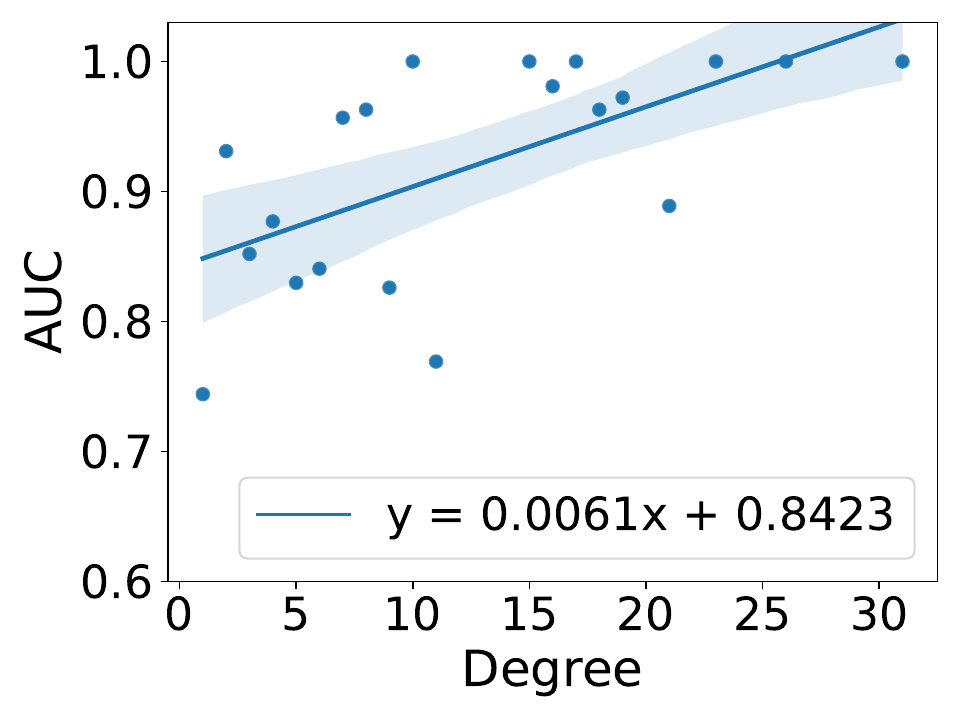}}
\subfloat[GRADATE on Cora]{\label{fig:bias3}\includegraphics[width=0.25\linewidth]{
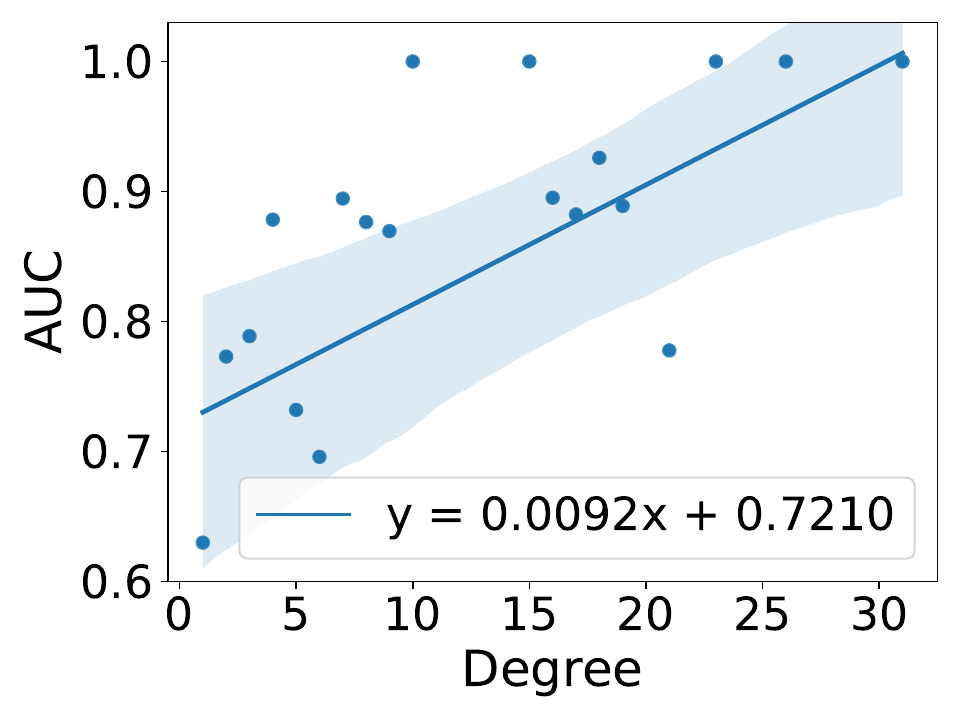}}
\subfloat[Our AD-GCL on Cora]{\label{fig:anal1}\includegraphics[width=0.25\linewidth]{
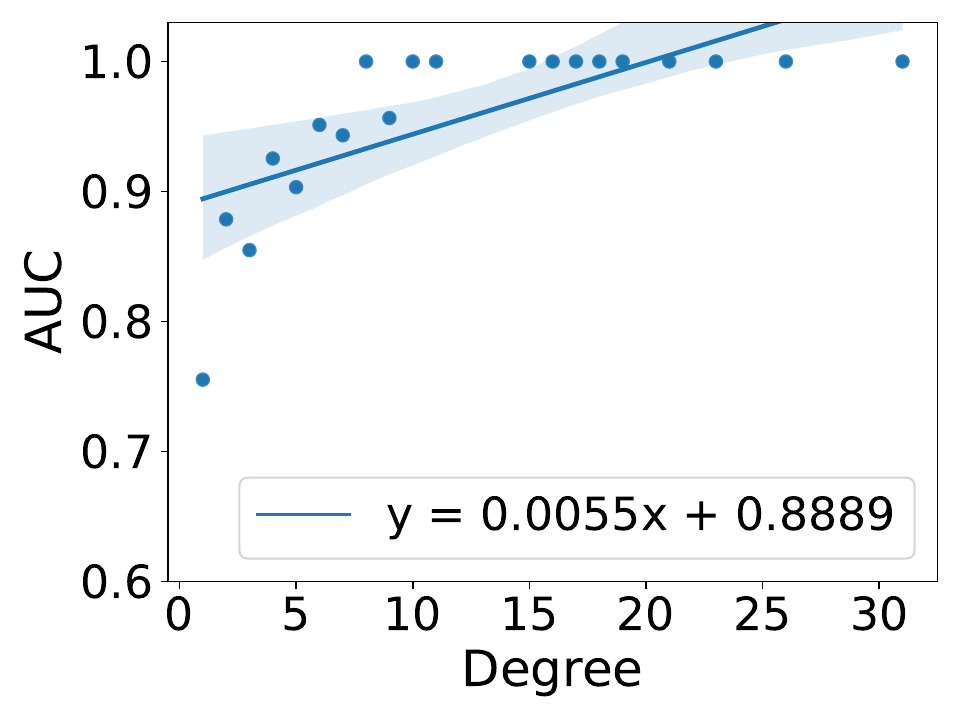}}\\	

\subfloat[CoLA on Citeseer]{\label{fig:bias5}\includegraphics[width=0.25\linewidth]{
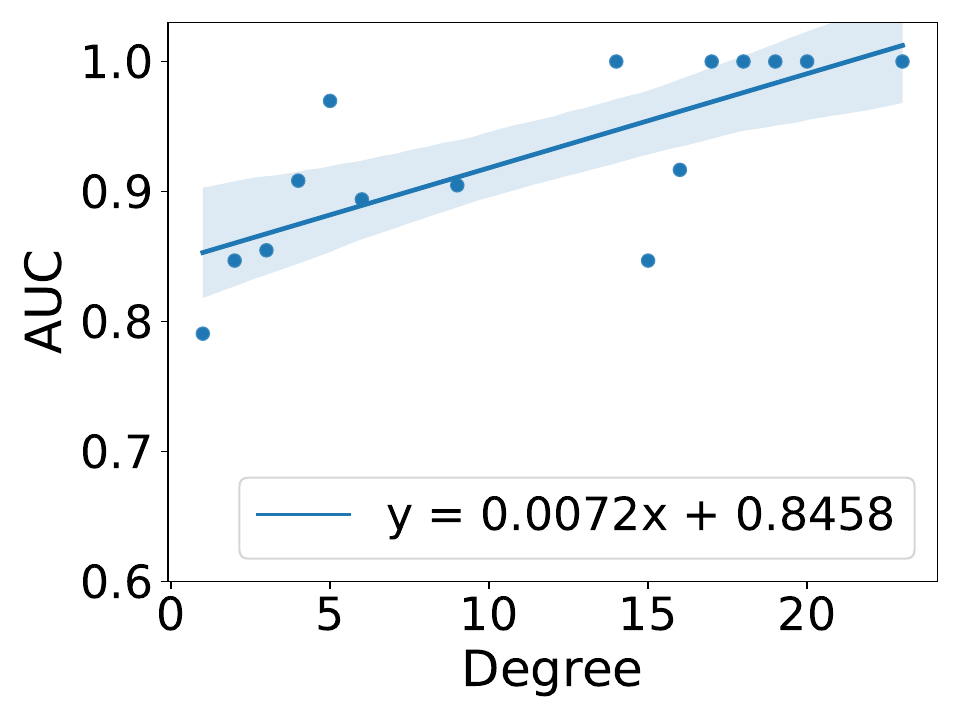}}
\subfloat[ANEMONE on Citeseer]{\label{fig:bias6}\includegraphics[width=0.25\linewidth]{
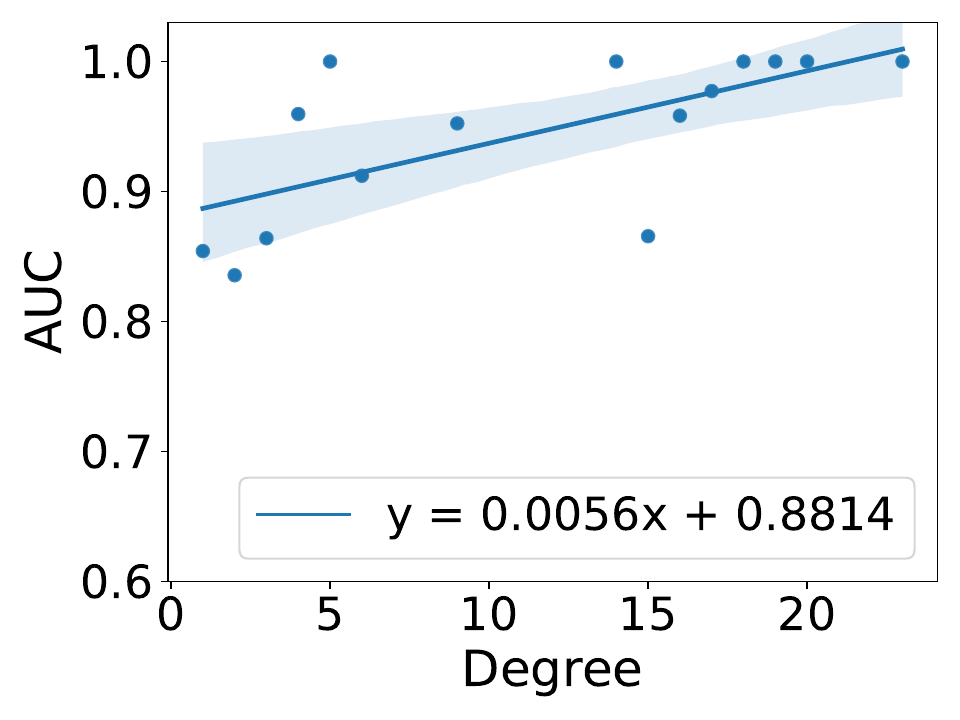}}
\subfloat[GRADATE on Citeseer]{\label{fig:bias7}\includegraphics[width=0.25\linewidth]{
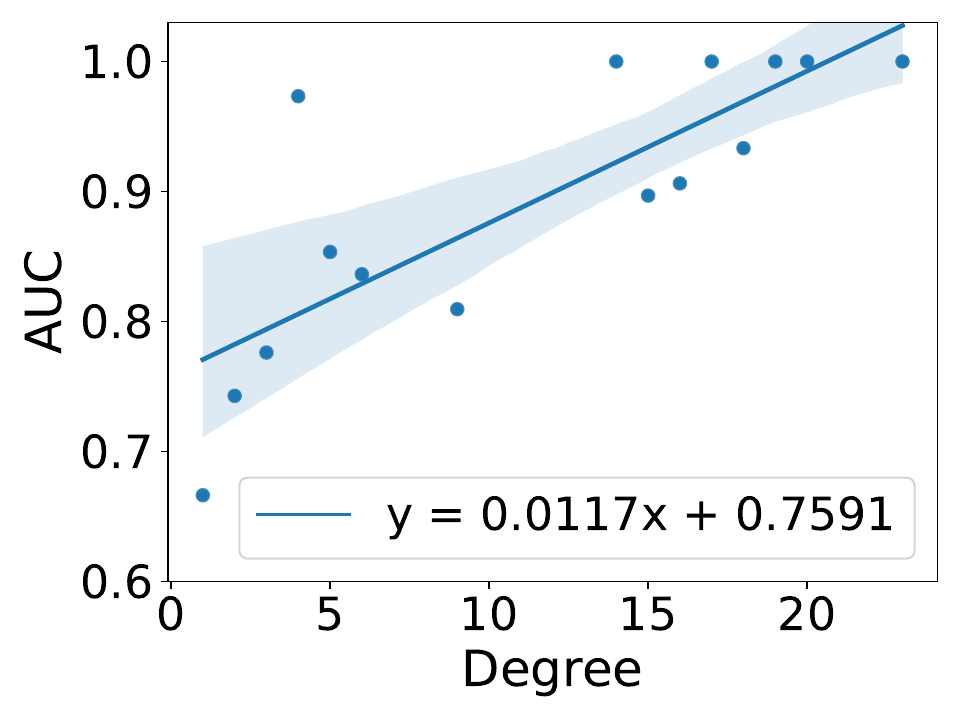}}
\subfloat[\small{Our AD-GCL on Citeseer}]{\label{fig:anal2}\includegraphics[width=0.25\linewidth]{
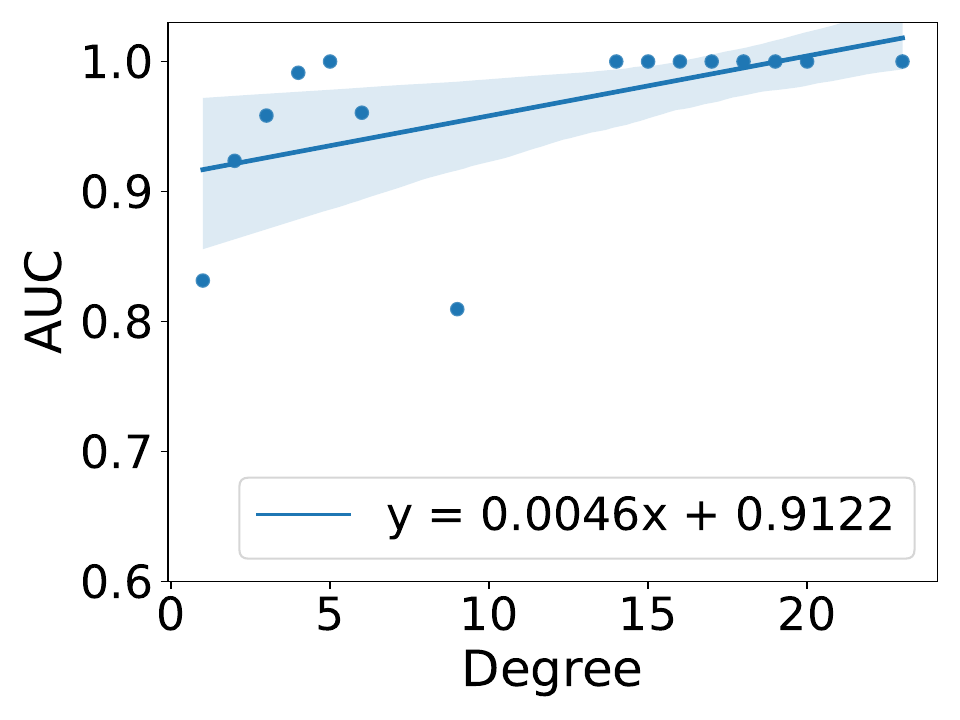}}\\	
\setlength{\belowcaptionskip}{-0.5cm}
\caption{An illustrative showcasing performance disparity between head and tail anomalies. Blue scatters represent the AUC at a specific degree. The blue line represents the regression line, with a steeper slope indicating a more pronounced structural bias. \looseness=-1}
\label{fig:bias}
\end{figure*}

Graph anomaly detection plays a crucial role in various domains including cybersecurity~\cite{ye2021fenet}, finance~\cite{zheng2023survey}, and distributed systems~\cite{pazho2023survey}. Its primary objective is to identify rare, unexpected, and suspicious observations that significantly deviate from the patterns of the majority in datasets, which helps minimize the considerable damage caused by detrimental events~\cite{ma2021comprehensive}. Traditional anomaly detection methods are mainly based on feature space to uncover anomalies, neglecting the relational information inherent in real-world data. On the other hand, labeling anomalous samples usually requires domain experts and is undoubtedly time-consuming and expensive. Fortunately, graph contrastive learning (GCL) has demonstrated its superiority in mining graph structural information from unlabeled data in the wild, which has led to extensive research on contrastive learning-based unsupervised graph anomaly detection (GAD) in recent years~\cite{liu2021anomaly,duan2023graph}.

Existing GAD techniques are predominantly designed to optimize overall detection performance, without paying much attention to the algorithm's robustness over structural imbalance~\cite{duan2023graph,chenboosting}. However, such structural imbalance commonly exists in most real-world graphs that follow the power-law degree distribution, where a substantial fraction of nodes have low degrees, termed as tail nodes while the rest are referred to as head nodes (see Appendix for statistical analysis on the degree distribution).
So it naturally raises a concern: can GCL-based anomaly detection models that combine the advantages of both graph neural networks (GNNs) and contrastive learning meet the challenge of structural imbalance? 
To figure this out, we choose three representative models for empirical study. As illustrated in Figure~\ref{fig:bias}, the detection performance notably drops as the node degree decreases. It reveals a nonnegligible finding that existing GCL-based anomaly detection methods struggle to effectively capture tail anomalies as we expected. This disparity has raised concerns regarding the security and robustness of current anomaly detection methods in identifying tail anomalies, which severely impair their application in several high-impact detection scenarios. For instance, in social networks, most identity impersonation attacks are not targeted at celebrities (usually with very high node degrees) but involve cloning ordinary individuals (with low degrees) within the network to engage in malicious activities~\cite{goga2015doppelganger,hooi2017graph}. Similarly, in transaction networks, fraudsters often establish multiple shell companies hidden within complex networks to commit one-time financial fraud and then shut down the business~\cite{pacini2018role}, resulting in massive low-degree anomalies. Motivated by this observation, a natural question arises: 
\textit{Can we improve the detection capability of tail anomalies without compromising the performance of head anomalies?} 

Through in-depth investigation, we find that the successful identification of anomalies within the contrastive learning paradigm is based on local inconsistency mining~\cite{chenboosting}. It begins with learning the matching patterns between normal nodes and their neighbors (i.e., satisfying the homophily assumption), which are dominant in the graph. In contrast, anomalies are consistently different from their local neighbors, and due to their irregularity and diversity, it is much more difficult to adapt to prevailing patterns in the data~\cite{liu2021anomaly,jin2021anemone,duan2023graph}. For example, CoLA~\cite{liu2021anomaly} enforces the discriminator to learn the neighbor matching patterns by treating nodes and their subgraphs sampled by a multi-round random walk with restart (RWR) as positive pairs, and subgraphs sampled based on other nodes as negative pairs. In the testing phase, the target node and its sampled neighboring structures are input into the optimized discriminator to derive a specific anomaly score. Notably, head anomalies possess more abundant structural links that ensure the diversity of constructed sample pairs, making it easier for the discriminator to model this difference. In contrast, the tail anomalies do not have enough structural relations to form sufficiently diverse contrast pairs, making the discriminator prone to overfit its sparse neighborhood and misinterpret these contrast pairs as normal graph patterns. Hence, the performance observed on tail anomalies declines. Inspired by the successful characteristics of head nodes, an intuitive solution is to enlarge the neighborhood of tail nodes by actively mining potential and underlying relational topologies to enhance the diversity of contrast samples and guide the discriminator to model the correct boundary between benign and anomaly. However, graph completion is not trivial due to the non-independently and identically distributed (i.i.d.) nature of the graph-structured data~\cite{wu2020graph}, naive completion may lead to suboptimal solutions by degrading the performance of neighboring nodes.

To improve the detection performance from the perspective of structural imbalance, especially the identification of tail anomalies, this paper presents a novel \textbf{A}nomaly \textbf{D}etection model based on \textbf{GCL}, named AD-GCL. Specifically, for tail nodes (classified based on topology only, regardless of whether the node is abnormal or not), we propose an anomaly-guided neighbor completion strategy to enlarge their limited neighboring receptive field, which employs a mixup between the ego network of the anchor tail node and the ego network of auxiliary nodes sampled based on anomaly scores and feature saliency information.
As for head nodes, we introduce a neighbor pruning strategy based on the saliency information of features between nodes. The advantages of this strategy are two-fold: (1) it effectively filters out noise information associated with head nodes, such as problematic edge relationships; (2) it forges head nodes as tail nodes and regards the original view and the neighbor pruning view as contrastive views. Based on this, the knowledge learned from head nodes is used as guidance to provide supervisory signals to the corresponding forged tail nodes, facilitating the detection of genuine tail nodes. By enforcing the discriminator to maximize the agreement between the original and augmented graphs in terms of anomaly scores and features of nodes, anomalies that deviate from the majority of the data will be assigned higher anomaly scores. Our main contributions are summarized as follows:

$\bullet$ \textbf{\textit{Significance}}: We investigate the limitations of existing GCL-based graph anomaly detection models from a structural imbalance perspective through detailed statistical analysis. Our findings disclose the undesirable performance of existing works in spotting tail anomalies, which severely limits their application in various high-impact detection scenarios in the open world.

$\bullet$ \textbf{\textit{Algorithm}}: 
We propose a novel algorithm AD-GCL, which introduces neighbor pruning to filter out noisy edges for head nodes and enhances the model's performance in detecting genuine tail nodes by aligning head nodes with forged tail nodes. Moreover, we implement neighbor completion to enlarge the neighborhood for tail nodes and incorporate intra- and inter-view consistency loss as self-supervision to uncover abnormal patterns. 

$\bullet$ \textbf{\textit{Experiments}}: Comprehensive performance evaluation of the whole, head, and tail nodes on six public datasets demonstrates the superiority of our proposed AD-GCL over the baselines. AD-GCL effectively enhances the detection quality for both tail and head anomalies.

\section{Related Work}
\textbf{Anomaly Detection on Graphs.} 
With the rise of graph-structured data, GAD has gained traction~\cite{xu2024learning}. DOMINANT~\cite{ding2019deep} measures feature and structure reconstruction errors. GAD-NR~\cite{roy2023gad} detects anomalies using reconstructed node neighborhoods. 
Contrastive learning, successful in various domains~\cite{he2020momentum,xu2023cldg}, has been extended to GAD. CoLA~\cite{liu2021anomaly} first introduces GCL by measuring whether the target node matches its neighbor sample pairs. ANEMONE~\cite{jin2021anemone} adds patch-level (i.e., node versus node) consistency, and GRADATE~\cite{duan2023graph}, Sub-CR~\cite{zhang2022reconstruction} and SAMCL~\cite{hu2023samcl} combine multi-view contrasts to estimate the anomaly score of nodes further. GADAM~\cite{chenboosting} designs adaptive message passing to avoid local anomaly signal loss. Despite their promise, GCL-based methods struggle with tail anomalies, reducing overall detection effectiveness.

\noindent\textbf{Imbalanced Learning on Graphs.} 
Imbalanced learning problems are widespread in real-world scenarios~\cite{liu2023survey}. Most research focuses on class imbalance~\cite{park2022graphens,zeng2023imgcl}, with some addressing it in GAD~\cite{liu2021pick}.
Subsequent work noted graph structural bias. Early approaches introduce degree-specific graph convolution~\cite{wu2019net,tang2020investigating}. Tail-GNN~\cite{liu2021tail} transfers neighborhood translations from head to tail nodes. GRADE~\cite{wang2022uncovering} shows the ability of GCL to handle structural imbalance. SAILOR~\cite{liao2023sailor} and GRACE~\cite{xu2023grace} enhance tail node representation by structural augmentation. Despite significant achievements, most methods are semi-supervised and focus on node classification. The ability of anomaly detection methods to handle structural imbalances has received less scrutiny.

\section{Methodology}
In this paper, we focus on the unsupervised anomaly detection problem for attributed graphs $\mathcal{G}= \left ( \mathbf{A},\mathbf{X} \right )$, where $\mathcal{V}=\left\{ v_{1},\cdots ,v_{n}\right\}$ is the set of nodes $\left ( \left| \mathcal{V}\right|=n \right ) $, $\mathbf{A}\in \mathbb{R}^{n\times n}$ denotes the adjacency matrix and $\mathbf{X}\in \mathbb{R}^{n\times f}$ is the attribute matrix. For each node $v\in \mathcal{V}$, $\mathcal{N}_{v}$ is the neighborhood of $v$ and $\left|\mathcal{N}_{v} \right| $ is the degree of $v$. Following the convention of prior works~\cite{liu2020towards,liu2021tail}, $K$ is predefined as a degree threshold, tail nodes are defined as $\mathcal{V}_{tail}=\left\{v:\left|\mathcal{N}_{v} \right|\leq K \right\}$, and other nodes are called head nodes $\mathcal{V}_{head}=\left\{u:\left|\mathcal{N}_{u} \right|> K \right\}$.

The goal of unsupervised anomaly detection is to learn an anomaly score function $f\left ( \cdot  \right ): \mathbb{R}^{n\times n}\times \mathbb{R}^{n\times d}\to\mathbb{R}^{n}$, which estimates the anomaly score $S_{i}=f\left ( v_{i} \right )$ of each node. The anomaly score $s_{i}$ measures the extent of abnormality for node $v_i$. A larger anomaly score means that it is more likely to be anomalous. Unlike previous anomaly detection work that only focused on the overall goal, in this paper, we pay attention to the robustness of structural imbalance, i.e., the detection performance of tail nodes $\left\{f\left ( v \right ):v\in \mathcal{V}_{tail} \right\}$ and head nodes $\left\{f\left ( u \right ):u\in \mathcal{V}_{head} \right\}$.

\begin{figure*}
  \centering
  \includegraphics[width=1\linewidth]{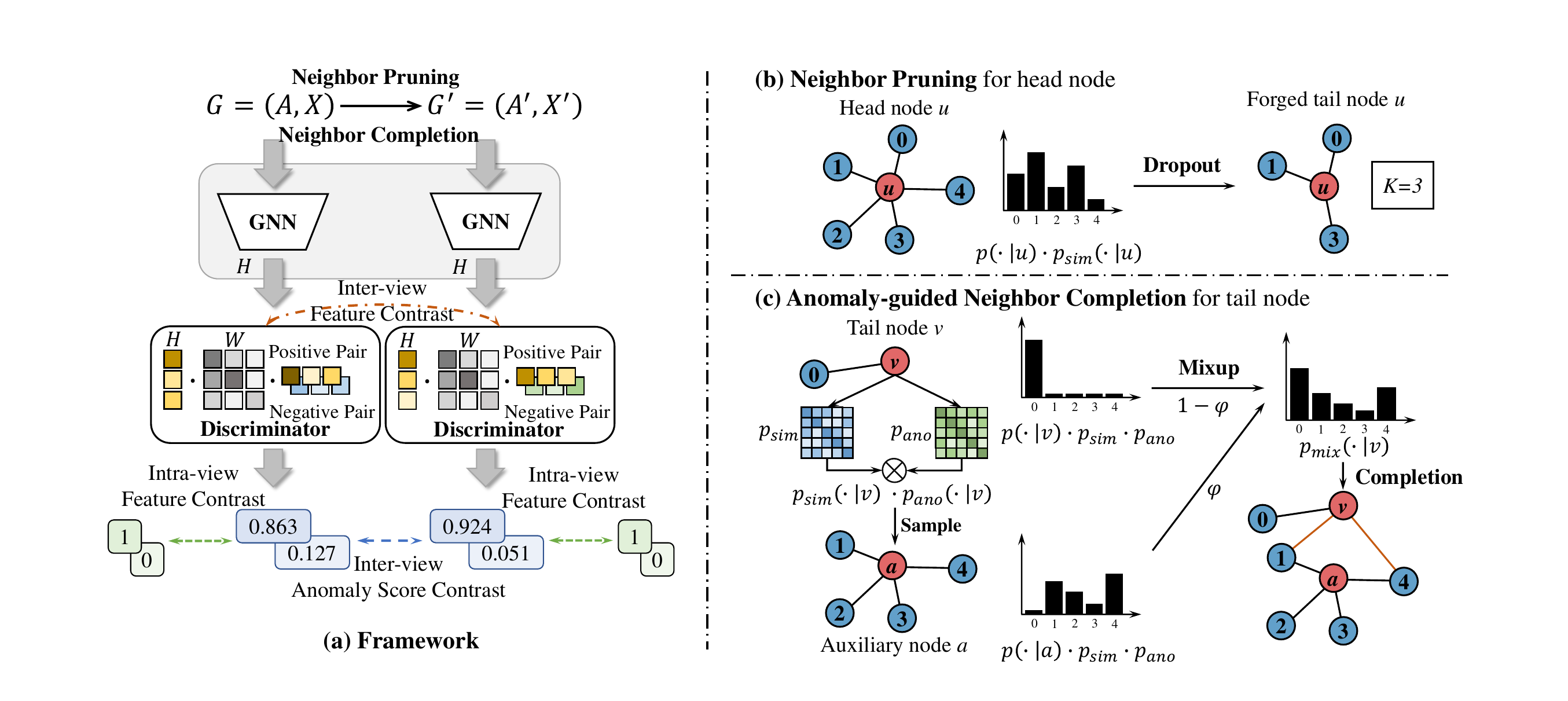}
  \caption{(a) The architecture of the AD-GCL; (b) Neighbor pruning to filter noise edges of head nodes; (c) Anomaly-guided neighbor completion to enlarge the receptive field of tail nodes. Intra- and inter-view contrast to enhance representation.}
  \label{fig:overview}
\end{figure*}

\subsection{Neighbor Pruning Strategy}
We propose a neighbor pruning strategy that leverages saliency information of inter-node features to effectively filter out noise information associated with head nodes, such as certain cunning abnormal objects that tend to establish relationships with these normal objects to camouflage themselves as conforming to the prevailing patterns~\cite{ma2021comprehensive}. Note that models often perform better on high-resource groups (i.e., head nodes) with sufficient structural information. Therefore, the neighbor pruning strategy forges head nodes as tail nodes, constructing contrastive pairs between the original view and the neighbor pruned view. We expect to utilize knowledge learned from head nodes to supervise forged tail nodes through alignment, thereby enhancing the model's performance on genuine tail nodes.
Specifically, for the head node $u\in \mathcal{V}_{head}$, we sample from the following multinomial distribution:
\begin{equation}
\label{eq:np}
\widetilde{\mathcal{N}}_{u} \sim \text{Multinomial}\left(K; p(\cdot | u) \cdot p_{\text{sim}}(\cdot | u)\right),
\end{equation}
where $ \left|\widetilde{\mathcal{N}}_{u} \right|=K$, so that the forged tail node $u$ retains $K$ important edges while discarding potential problematic or redundant relationships in the head nodes. The neighbor distribution $p\left ( v |u \right )=1/\left|\mathcal{N}_{u} \right|$ if $v\in \mathcal{N}_{u}$ and $p\left ( v |u \right )=0$ otherwise. 
The feature similarity distribution $p_{sim}\left ( v |u \right )=\textrm{sim} \left ( x_{v},x_{u} \right )$ for $v\neq u$ and $p_{sim}\left ( u |u \right )=0$, where all the features are $\ell_2$  normalized and dot product (cosine) similarity is used to compare them $\mathrm{sim}(q, k) = q^\top k / \lVert q\rVert \lVert k\rVert$.

Unlike existing strategies based on supervised label-aware~\cite{dou2020enhancing} or degree-proportional pruning~\cite{wang2022uncovering}, we introduce a similarity-aware method to prune head nodes into forged tail nodes instead of discarding them proportionally. By employing self-supervised alignment between head and forged tail nodes in both features and anomaly scores, we enhance the discriminator's ability to model accurate decision boundaries for both node types (refer to the ablation study).

\subsection{Anomaly-Guided Neighbor Completion} \label{sec:nc}
To improve the detection performance of tail nodes, a natural solution is to enlarge the tail node neighborhood by complementing potential or hidden relationships. However, due to the non-i.i.d. nature of the graph, biased graph completion may lead to a further decline in performance (as shown in Appendix).
To address the aforementioned problem, we propose a novel anomaly-guided neighbor completion strategy for tail nodes. We first adopt a joint distribution of anomaly score similarity and feature saliency to sample auxiliary nodes similar to anchor tail nodes. Subsequently, we enlarge the neighborhood of tail nodes by mixing the ego network of anchor tail nodes and the ego network of auxiliary nodes. Different from PC-GNN~\cite{liu2021pick}, which incorporates label distribution information to design a neighbor sampler for over-sampling the neighborhood of the minority class nodes, our method effectively enhances the receptive field of tail nodes in an unsupervised manner by integrating both anomaly and feature information.

In specific, we first derive a scoring matrix $S$ to record the abnormality of nodes.  Let $t$ represent the current number of training epochs. In each epoch, the discriminator generates anomaly scores $s_{v,p}^{t}$ and $s_{v,n}^{t}$ for the anchor tail node with its positive and negative pairs, respectively (refer to Section Graph Contrastive Network for detailed explanation). Considering the fluctuation of anomaly scores during training, we use a sliding window to store the training results of the past $w$ epochs. For a specific tail node $v$, its anomaly score is represented as $S_{v}=\left [ s_{v,p}^{t-w+1},s_{v,n}^{t-w+1},\cdots , s_{v,p}^{t},s_{v,n}^{t}\right ] $ , and $S \in \mathbb{R}^{n\times 2w}$. The abnormal similarity distribution is modeled as $p_{ano}\left ( u,v \right )=S_{u}\cdot S_{v}^T$. For any tail node $v$, we jointly consider anomaly score similarity and feature similarity, and sample an auxiliary node $a$ from the multimodal distribution $\text{Multinomial}\left(p_{nc}\left ( \cdot |v \right )\right)$, where $p_{nc}\left ( \cdot |v \right ) = p_{\text{sim}}(\cdot | v) \cdot p_{\text{ano}}(\cdot | v)$. Then, we form a larger neighborhood distribution by mixing up the ego network of nodes $v$ and $a$. The neighborhood distribution of $v$ after mixup is defined as:
\begin{equation}
\resizebox{1\hsize}{!}{$
p_{mix} \left ( \cdot |v \right )=\left ( 1- \varphi \right )\left ( p\left ( \cdot |v \right )\cdot p_{nc}\left ( \cdot |v \right ) \right )+\varphi \left ( p\left ( \cdot |a \right )\cdot p_{nc}\left ( \cdot |a \right ) \right ),
$}
\end{equation}
where the mixing ratio $\varphi$ of auxiliary node $a$ increases with the similarity to the tail node $v$, and $\varphi$ is at most 0.5 to preserve the original neighborhood of the anchor tail node. Finally, we sample neighbors for the tail node $v$ from the mixup neighborhood distribution $p_{mix} \left ( \cdot |v \right )$. The number of neighbors is sampled from the degree distribution of the original graph to keep degree statistics.

\subsection{Graph Contrastive Network} \label{sec:gcn}
The key challenge in contrastive learning is selecting appropriate contrastive pairs. Most normal nodes tend to adhere to the homophily hypothesis, exhibiting potential matching patterns with their neighbors. On the contrary, anomalous nodes often demonstrate inconsistencies with their neighbors. Based on this, the graph contrastive network treats the anchor node and its neighbors as intra-view positive pairs, and the neighbors of other nodes as intra-view negative pairs. Furthermore, during the initial training stage, we regard the original graph (called $view 1$) and the neighbor pruning graph (called $view 2$) as contrastive views, and utilize well-trained head nodes from the original graph to supervise the corresponding forged tail nodes, i.e., the corresponding nodes are inter-view positive pairs.
In the later stage of training, we use the neighbor completion strategy to enlarge the receptive field of tail nodes, generating two augmented graphs (referred to as $view 1$ and $view 2$) as contrastive views to further enhance node representations. \looseness=-1

Specifically, given a graph $\mathcal{G}^{view}= \left ( \mathbf{A}^{view},\mathbf{X}^{view} \right )$, where $view$ denote contrastive view 1 or 2, representing the original graph or augmentation graph. The node representations are updated by a GNN-based encoder $f\left ( \cdot  \right )$: 
\begin{equation}
\label{eq:gnn}
\mathbf{H}^{view} = f\left ( \mathbf{A}^{view},\mathbf{X}^{view} \right ),
\end{equation}
where $\mathbf{H}^{view}\in \mathbb{R}^{n\times d}$ are the node representations of graph $\mathcal{G}^{view}$. 

Then, we use the \textrm{Readout} function to calculate the neighbor representation of the node, which has been extensively used in prior works~\cite{xu2023cldg}. The formal expression is as follows:
\begin{equation}
\label{eq:readout}
\mathbf{h}_{\mathcal{N}_{i}}^{view}= \textrm{Readout} \left ( \left\{ \mathbf{h}_{j}^{view},\forall j\in \hat{\mathcal{N}}_{i}^{view}\right\} \right ),
\end{equation}
where $\mathbf{h}_{\mathcal{N}_{i}}^{view}\in \mathbb{R}^{d}$ is the neighbor representation of node $i$. $\hat{\mathcal{N}}_{i}^{view}$ represents the neighbor set sampled for node $i$ using random walk with restart (RWR)~\cite{tong2006fast} in $\mathcal{G}^{view}$. We adopt the average pooling function as our readout function. This approach is simple and effective, while not introducing additional model parameters.

Inspired by existing anomaly detection works based on GCL~\cite{liu2021anomaly}, we devise a discriminator module to evaluate the potential edge relationships in the graph, leveraging both node representations and node neighbor representations. The discriminator is built by the bilinear scoring function. The predicted similarity score of the discriminator can be calculated as follows:
\begin{equation}
s_{i}^{view}=\textrm{Bilinear}\left ( \mathbf{h}_{\mathcal{N}_{j}}^{view}, \mathbf{h}_{i}^{view} \right )=\sigma \left ( \mathbf{h}_{\mathcal{N}_{j}}^{view}\mathbf{W} {\mathbf{h}_{i}^{view}}^{\top } \right ),
\end{equation}
where $\mathbf{W}$ is a trainable matrix, and $\sigma\left ( \cdot  \right ) $ is $\textrm{Sigmoid}$ function. When the input to the discriminator is the potential neighbor relationship in the graph, i.e., $i= j$, the output score of the positive pair is $s_{i,p}^{view}$. When the input to the discriminator is a negative pair, i.e., $i\neq j$, the output score is $s_{i,n}^{view}$. $s_{v,p}^{t}$ are mentioned in the previous subsection is $s_{v,p}^{t}=s_{v,p}^{view1} + s_{v,p}^{view2}$ at training step $t$.

The representations of normal nodes, which are dominant in the graph, and their neighbors tend to be similar, aiming for $s_{i,p}^{view}$ to be close to 1. In contrast, normal nodes and neighbors of other nodes may be dissimilar in negative pairs. Therefore, we adopt binary cross-entropy (BCE) loss~\cite{velivckovic2018deep} to train the intra-view contrastive pairs:
\begin{equation}
\mathcal{L}_{intra}^{view}=-\frac{1}{2 n} \sum_{i=1}^n\left(\log \left(s_{i,p}^{view}\right)+\log \left(1-s_{i,n}^{view}\right)\right),
\end{equation}
where $\mathcal{L}_{intra}$ is the sum of losses from both views, denoted as $\mathcal{L}_{intra} = \mathcal{L}_{intra}^{view1} + \mathcal{L}_{intra}^{view2}$. 

Then, we enforce the agreement between the two contrastive views by InfoNCE~\cite{van2018representation}. We treat the features learned and scores output by the discriminator of the corresponding nodes as inter-view positive pairs. The inter-view contrastive loss is as follows:
\begin{equation}
\mathcal{L}_{inter} = \textrm{ctr}\left ( \mathbf{H}^{view1},\mathbf{H}^{view2} \right )+\textrm{ctr}\left ( \mathbf{S}^{view1},\mathbf{S}^{view2} \right )
\end{equation}
where $\mathbf{H}$ represents the node feature and $\mathbf{S}$ is the score output by the discriminator. $\textrm{ctr}\left ( \cdot ,\cdot  \right )$ is as follows:
\begin{equation}
\textrm{ctr}\left (\mathbf{Q},\mathbf{K} \right )=-\sum_{i=1}^{N}\log\frac{\exp{\left ( \mathbf{q}_{i}\cdot \mathbf{k}_{i}/\tau  \right )}}{\sum_{j=1}^{N}\exp{\left ( \mathbf{q}_{i}\cdot \mathbf{k}_{j}/\tau  \right )}},
\end{equation}
where $\tau$ is a temperature parameter. 

In the training phase, we jointly train the intra-view and inter-view contrastive losses. The overall objective is defined as follows:
\begin{equation}
\label{eq:loss}
\mathcal{L}=\mathcal{L}_{intra} + \alpha \mathcal{L}_{inter}
\end{equation}
where $\alpha$ is a trade-off parameter to balance the importance between two objectives. Detailed complexity analysis is given in the Appendix.

\subsection{Anomaly Score Calculation}
After the graph contrastive network is well-trained, we use a statistical anomaly estimator~\cite{liu2021anomaly} during the inference stage to calculate anomaly scores for each node based on local inconsistency. For node $i$, the statistical anomaly estimator generates the final anomaly score $S_i$ through multiple rounds of positive and negative sampling prediction scores:
\begin{equation}
\label{eq:asc}
S_{i}=\frac{\sum_{r=1}^{R}\left ( s_{i,n}^{r}-s_{i,p}^{r} \right )}{R}
\end{equation}
where $R$ is the number of sampling rounds. 
The scores of the negative pair $s_{i,n}^{r}$ and positive pair $s_{i,p}^{r}$ are obtained from the discriminator in the $r$-th round. For a normal node, the positive pair score $s_{i,p}$ should be close to 1, while the negative pair score $s_{i,n}$ should be close to 0. However, the discriminator struggles to distinguish the matching patterns of positive and negative pairs for abnormal nodes. The higher the score of $S_i$, the stronger the indication of abnormality.

\section{Experiments} \label{sec:sec5}

\begin{table*}
\setlength{\belowcaptionskip}{-0.3cm}
\renewcommand\arraystretch{1.075} 
\resizebox{1\textwidth}{!}{
\centering
\begin{tabular}{cccccccccccc}
\toprule \hline
                              & \multirow{2}{*}{Method} & \multicolumn{3}{c}{Cora} & \multicolumn{3}{c}{Citeseer} & \multicolumn{3}{c}{Pubmed}  \\
                              \cline{3-11}
                              &                         & AUC     & TN     & HN        & AUC        & TN     & HN    & AUC      & TN      & HN       \\ \hline
    & SCAN   
& 67.42$_{\pm 0.14}$ & 46.74$_{\pm 0.27}$ & 74.22$_{\pm 0.18}$ & 69.80$_{\pm 0.74}$ & 48.70$_{\pm 0.84}$ & 76.81$_{\pm 0.14}$ & 74.16$_{\pm 0.68}$ & 49.77$_{\pm 0.38}$ & 89.93$_{\pm 0.06}$ \\ 
    & Radar      
& 52.62$_{\pm 0.33}$ & 48.89$_{\pm 0.79}$ & 54.39$_{\pm 0.46}$ & 62.30$_{\pm 0.87}$ & 49.32$_{\pm 0.96}$ & 58.86$_{\pm 0.04}$ & 49.34$_{\pm 0.49}$ & 49.75$_{\pm 0.01}$ & 50.31$_{\pm 0.30}$\\ 
    & ONE    
& 49.20$_{\pm 0.42}$ & 48.87$_{\pm 0.44}$ & 49.52$_{\pm 0.78}$ & 49.79$_{\pm 0.10}$ & 49.66$_{\pm 0.13}$ & 50.09$_{\pm 0.91}$ & 50.28$_{\pm 0.82}$ & 50.02$_{\pm 0.78}$ & 50.34$_{\pm 0.52}$ \\ 
\hline
    & AEGIS    
& 48.55$_{\pm 1.34}$ & 46.71$_{\pm 1.40}$ & 50.29$_{\pm 1.41}$ & 47.67$_{\pm 1.05}$ & 47.66$_{\pm 1.26}$ & 48.57$_{\pm 1.70}$ & 57.02$_{\pm 1.24}$ & 64.51$_{\pm 1.37}$ & 50.93$_{\pm 1.85}$ \\ 
    & GAAN      
& 53.54$_{\pm 1.29}$ & 50.69$_{\pm 1.49}$ & 53.20$_{\pm 1.80}$ & 51.48$_{\pm 1.44}$ & 49.92$_{\pm 1.27}$ & 50.24$_{\pm 1.08}$ & 52.66$_{\pm 1.49}$ & 48.94$_{\pm 1.03}$ & 51.45$_{\pm 1.14}$ \\ 
    & CoLA                  
& 86.45$_{\pm 0.62}$ & 82.59$_{\pm 0.36}$ & 87.88$_{\pm 0.43}$ & 89.04$_{\pm 0.56}$ & 82.31$_{\pm 0.77}$ & $\underline{97.71_{\pm 0.35}}$ & 93.54$_{\pm 0.37}$ & 92.38$_{\pm 0.50}$ & 97.08$_{\pm 0.29}$\\ 
    & ANEMONE                   
& $\underline{90.70_{\pm 0.91}}$ & $\underline{85.09_{\pm 0.29}}$ & $\underline{94.19_{\pm 0.61}}$ & $\underline{92.04_{\pm 0.38}}$ & $\underline{87.83_{\pm 0.91}}$ & 96.38$_{\pm 0.55}$ & $\underline{95.29_{\pm 0.13}}$ & $\underline{93.96_{\pm 0.19}}$ & $\mathbf{98.20}_{\pm 0.11}$ \\ 
    & AdONE                     
& 67.68$_{\pm 0.48}$ & 33.73$_{\pm 1.03}$ & 80.00$_{\pm 1.00}$ & 69.68$_{\pm 0.22}$ & 43.61$_{\pm 0.34}$ & 91.44$_{\pm 0.36}$ & 86.54$_{\pm 0.16}$ & 85.37$_{\pm 1.13}$ & 73.49$_{\pm 1.03}$ \\ 
    & GRADATE                  
& 86.99$_{\pm 0.74}$ & 77.87$_{\pm 0.79}$ & 91.56$_{\pm 0.18}$ & 81.71$_{\pm 0.81}$ & 72.17$_{\pm 1.08}$ & 93.54$_{\pm 1.50}$ & 87.44$_{\pm 1.08}$ & 79.46$_{\pm 0.67}$ & 97.33$_{\pm 0.35}$ \\
    & GAD-NR                    
& 70.83$_{\pm 0.56}$ & 46.06$_{\pm 0.30}$ & 72.24$_{\pm 0.36}$ & 73.28$_{\pm 0.15}$ & 49.91$_{\pm 0.30}$ & 86.04$_{\pm 0.39}$ & 71.25$_{\pm 0.16}$ & 49.52$_{\pm 0.77}$ & 68.81$_{\pm 0.82}$ \\

\hline                        
    & AD-GCL                                    
& $\mathbf{92.83}_{\pm 0.43}$     & $\mathbf{85.70}_{\pm 1.12}$        & $\mathbf{98.67}_{\pm 0.12}$           & $\mathbf{94.88}_{\pm 0.27}$     & $\mathbf{90.51}_{\pm 0.56}$       & $\mathbf{99.71}_{\pm 0.56}$      & $\mathbf{95.74}_{\pm 0.10}$         & $\mathbf{95.12}_{\pm 0.25}$       & $\underline{97.88_{\pm 0.05}}$  \\

\hline\bottomrule
\end{tabular}}


\resizebox{1\textwidth}{!}{
\centering
\begin{tabular}{cccccccccccc}

                              & \multirow{2}{*}{Method} & \multicolumn{3}{c}{Bitcoinotc} & \multicolumn{3}{c}{BITotc} & \multicolumn{3}{c}{BITalpha} \\
                              \cline{3-11}
                              &                         & AUC     & TN     & HN        & AUC        & TN     & HN    & AUC      & TN      & HN       \\ \hline
    & SCAN   
& 66.37$_{\pm 0.64}$ & 51.64$_{\pm 0.96}$ & 69.64$_{\pm 0.02}$ & 64.92$_{\pm 0.76}$ & 47.59$_{\pm 0.10}$ & 70.02$_{\pm 0.91}$ & 66.28$_{\pm 0.93}$ & 49.39$_{\pm 0.23}$ & 72.30$_{\pm 0.02}$ \\
    & Radar      
& 47.34$_{\pm 0.27}$ & 44.61$_{\pm 0.17}$ & 48.64$_{\pm 0.02}$ & 49.96$_{\pm 0.09}$ & 48.44$_{\pm 0.11}$ & 52.48$_{\pm 0.07}$ & 53.09$_{\pm 0.42}$ & 52.91$_{\pm 0.92}$ & 53.73$_{\pm 0.85}$\\
    & ONE    
& 49.71$_{\pm 0.72}$ & 48.95$_{\pm 0.10}$ & 50.11$_{\pm 0.05}$ & 49.55$_{\pm 0.04}$ & 48.88$_{\pm 0.10}$ & 50.13$_{\pm 0.32}$ & 49.64$_{\pm 0.47}$ & 49.07$_{\pm 0.50}$ & 50.13$_{\pm 0.05}$\\
\hline
    & AEGIS    
& 52.79$_{\pm 1.06}$ & 57.29$_{\pm 1.18}$ & 50.46$_{\pm 1.60}$ & 53.64$_{\pm 1.24}$ & 56.41$_{\pm 1.92}$ & 52.62$_{\pm 1.72}$ & 54.21$_{\pm 1.11}$ & 57.02$_{\pm 1.36}$ & 51.14$_{\pm 1.30}$ \\
    & GAAN      
& 58.32$_{\pm 1.32}$ & 49.80$_{\pm 1.88}$ & 53.34$_{\pm 1.72}$ & 54.52$_{\pm 1.95}$ & 49.75$_{\pm 1.37}$ & 50.71$_{\pm 1.82}$ & 55.62$_{\pm 1.82}$ & 50.39$_{\pm 1.85}$ & 51.69$_{\pm 1.91}$\\
    & CoLA                  
& $\underline{79.30_{\pm 0.42}}$ & 65.00$_{\pm 0.23}$ & 70.13$_{\pm 0.99}$ & 80.11$_{\pm 0.74}$ & 66.42$_{\pm 0.92}$ & 85.41$_{\pm 0.83}$ & $\underline{76.91_{\pm 0.29}}$ & $\underline{66.62_{\pm 0.66}}$ & 74.28$_{\pm 0.99}$\\
    & ANEMONE                   
& 79.19$_{\pm 1.13}$ & 65.00$_{\pm 0.95}$ & $\underline{84.92_{\pm 0.68}}$ & $\underline{80.90_{\pm 0.42}}$ & 68.37$_{\pm 0.25}$ & $\underline{85.67_{\pm 0.84}}$ & 75.44$_{\pm 0.88}$ & 64.44$_{\pm 0.27}$ & 72.73$_{\pm 1.09}$\\
    & AdONE
& 79.24$_{\pm 0.68}$ & $\underline{68.64_{\pm 0.19}}$ & 75.72$_{\pm 0.41}$ & 80.57$_{\pm 0.30}$ & $\mathbf{72.06}_{\pm 0.86}$ & 76.88$_{\pm 0.39}$ & 76.34$_{\pm 0.29}$ & 60.52$_{\pm 0.30}$ & $\underline{79.34_{\pm 1.06}}$\\
    & GRADATE
& 73.70$_{\pm 0.98}$ & 49.38$_{\pm 0.21}$ & 84.46$_{\pm 0.32}$ & 74.01$_{\pm 0.22}$ & 50.93$_{\pm 0.88}$ & 82.55$_{\pm 0.42}$ & 69.34$_{\pm 0.97}$ & 51.89$_{\pm 0.65}$ & 68.09$_{\pm 0.63}$\\
    & GAD-NR                    
& 69.82$_{\pm 0.29}$ & 48.53$_{\pm 0.88}$ & 61.55$_{\pm 0.78}$ & 69.78$_{\pm 0.85}$ & 52.77$_{\pm 0.77}$ & 60.08$_{\pm 0.45}$ & 71.76$_{\pm 0.86}$ & 51.21$_{\pm 0.99}$ & 67.26$_{\pm 0.79}$\\
\hline                              
    & AD-GCL                                    
& $\mathbf{82.19}_{\pm 0.46}$       & $\mathbf{69.18}_{\pm 0.95}$       & $\mathbf{87.53}_{\pm 0.53}$& $\mathbf{82.11}_{\pm 0.20}$      & $\underline{70.39_{\pm 0.48}}$      & $\mathbf{86.25}_{\pm 0.20}$      & $\mathbf{79.62}_{\pm 0.66}$      & $\mathbf{67.73}_{\pm 1.04}$      & $\mathbf{82.77}_{\pm 1.29}$\\ 
\hline  \bottomrule
\end{tabular}}
\caption{Performance comparison for AUC (\%). TN and HN represent the AUC for tail nodes and head nodes, respectively. Bold indicates the optimal and underline indicates the suboptimal.}
\label{tab:auc}
\end{table*}

\subsection{Experiment Settings}
\textbf{Datasets.}\label{dataset} To conduct a comprehensive comparison, we evaluate AD-GCL on six widely used benchmark datasets for anomaly detection. Specifically, we choose two categories of datasets: 1) citation networks~\cite{liu2021anomaly,jin2021anemone} including  Cora, Citeseer, and Pubmed, 2) bitcoin trading networks~\cite{kumar2016edge, kumar2018rev2} including  Bitcoinotc, BITotc, and BITalpha. 

\noindent\textbf{Baselines.} We compare AD-GCL with ten popular methods for anomaly detection: three classical non-deep methods, i.e., clustering-based SCAN~\cite{xu2007scan}, matrix factorization-based Radar~\cite{li2017radar} and ONE~\cite{bandyopadhyay2019outlier}; seven neural network based frameworks, i.e., AEGIS~\cite{ding2021inductive}, GAAN~\cite{chen2020generative}, CoLA~\cite{liu2021anomaly}, ANEMONE~\cite{jin2021anemone}, AdONE~\cite{bandyopadhyay2020outlier}, GRADATE~\cite{duan2023graph}, and GAD-NR~\cite{roy2023gad}.

\noindent\textbf{Evaluation Metrics.} For evaluating the performance of the proposed methods, we employ a widely-used anomaly detection metric ROC-AUC~\cite{chen2020generative,ding2021inductive,duan2023graph,jin2021anemone,li2017radar,liu2021anomaly,roy2023gad}. 
The ROC curve plots the true positive rate against the false positive rate, while AUC is the area under the ROC curve with a higher value indicating better performance.
We also report the AUPRC and AP results in the Appendix.
To comprehensively analyze the anomaly detection performance of the model from the structural imbalance perspective, we adhere to the convention of previous research~\cite{liu2020towards,liu2021tail} by dividing head and tail nodes based on a predefined degree threshold $K$ and reporting the corresponding results. 
Please refer to the Appendix for more detailed parameter settings and experimental environment introduction. \looseness=-1

\subsection{Main Results and Analysis}
We evaluate the anomaly detection performance of AD-GCL by conducting a comparison with 10 baseline methods. By calculating the area under the ROC curve, the AUC scores of the 6 benchmark datasets are compared as shown in Table~\ref{tab:auc}. More experiments validating the effectiveness of AD-GCL are presented in Appendix, including report results for AUPRC and AP, comparisons of ROC curves, visualization of the distribution of abnormal scores, compared with naive data augmentation methods, the advantages of polynomial distribution sampling, and the sensitivity analysis of the readout function. Based on the results, we have the following observations: 

We can intuitively observe that AD-GCL achieves the best anomaly detection performance on these six datasets. It outperforms the runner-up methods by 2.13\%, 2.84\%, 0.45\%, 2.89\%, 1.21\%, and 2.71\%, respectively. Of greater significance, AD-GCL achieves the best AUC scores for both tail nodes and head nodes on most of the datasets.
Clustering method SCAN, matrix factorization method Radar and ONE show unsatisfactory performance compared to deep GAD methods. SCAN shows significant differences in the performance of head and tail nodes. Non-deep methods have inherent limitations in dealing with complex graph structures and high-dimensional and sparse features. Meanwhile, in datasets with sparse structures and features, such as Cora and Citeseer, the reconstruction quality of AdONE based on autoencoder methods tends to be poor, leading to completely collapsed solutions and consequently resulting in degraded performance. \looseness=-1

Within the category of deep anomaly detection methods, we observe that GCL-based approaches CoLA, ANEMONE, and GRADATE exhibit superior performance. This suggests that the GCL-based paradigm effectively detects anomalies by leveraging the feature and structure information present in the graph.
However, existing GCL-based methods focus on how to design contrast pairs of different granularities in graphs, such as node-subgraph or node-node, and they neglect to consider GAD from the perspective of structural imbalance. We first show the performance gap caused by structural imbalance through empirical research, and propose different strategies for imbalanced structures. Compared with existing GCL-based methods in Fig. ~\ref{fig:bias}, AD-GCL demonstrates significant performance improvements on nodes with varying degrees in Cora and Citeseer datasets.
We further validate that AD-GCL assigns higher anomaly scores to anomaly nodes via violin plots, and the experiments shown in Appendix. Our method yields surprising results by not only enhancing the performance of tail nodes but also improving the performance of head nodes. This highlights the significant role played by the proposed scheme.

\begin{figure*}[]
\centering
\setlength{\belowcaptionskip}{-0.3cm}
\includegraphics[width=1\linewidth]{
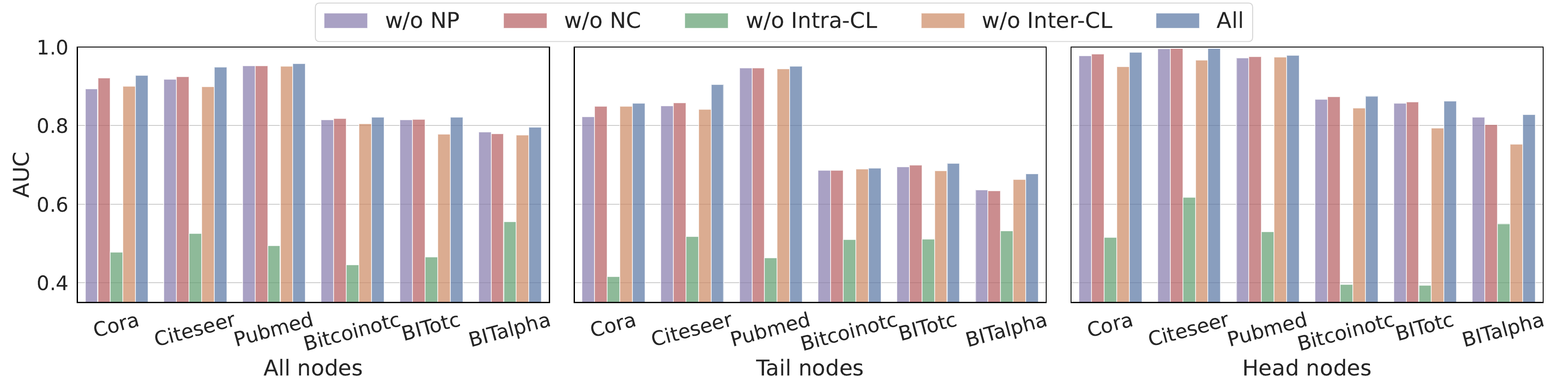}
\caption{Ablation study on different variants.}
\label{fig:ablation}    
\end{figure*}

\begin{figure*}[]
\centering
\setlength{\belowcaptionskip}{-0.3cm}
\subfloat[Sampling rounds $R$]{\label{fig:param1}\includegraphics[width=0.25\linewidth]{
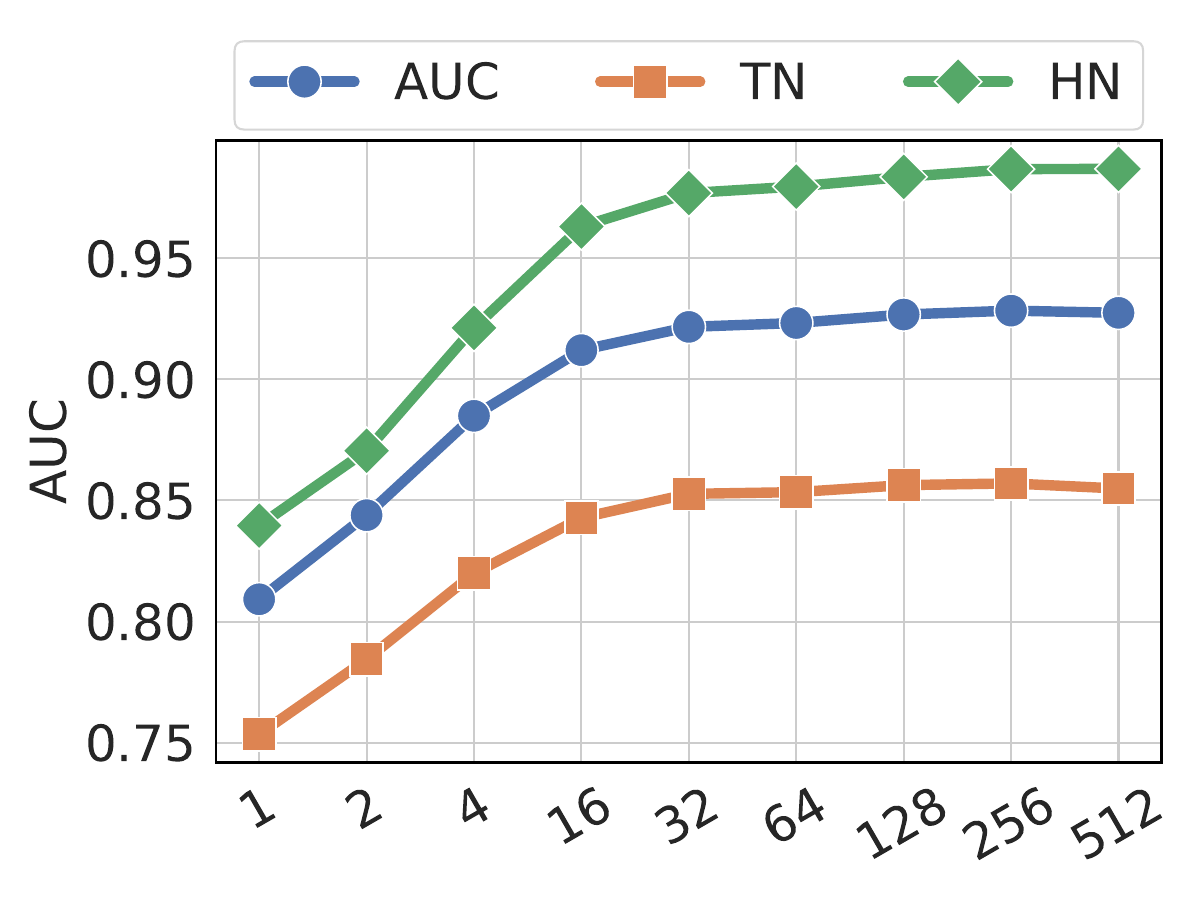}}
\subfloat[Dimension $d$]{\label{fig:param2}\includegraphics[width=0.25\linewidth]{
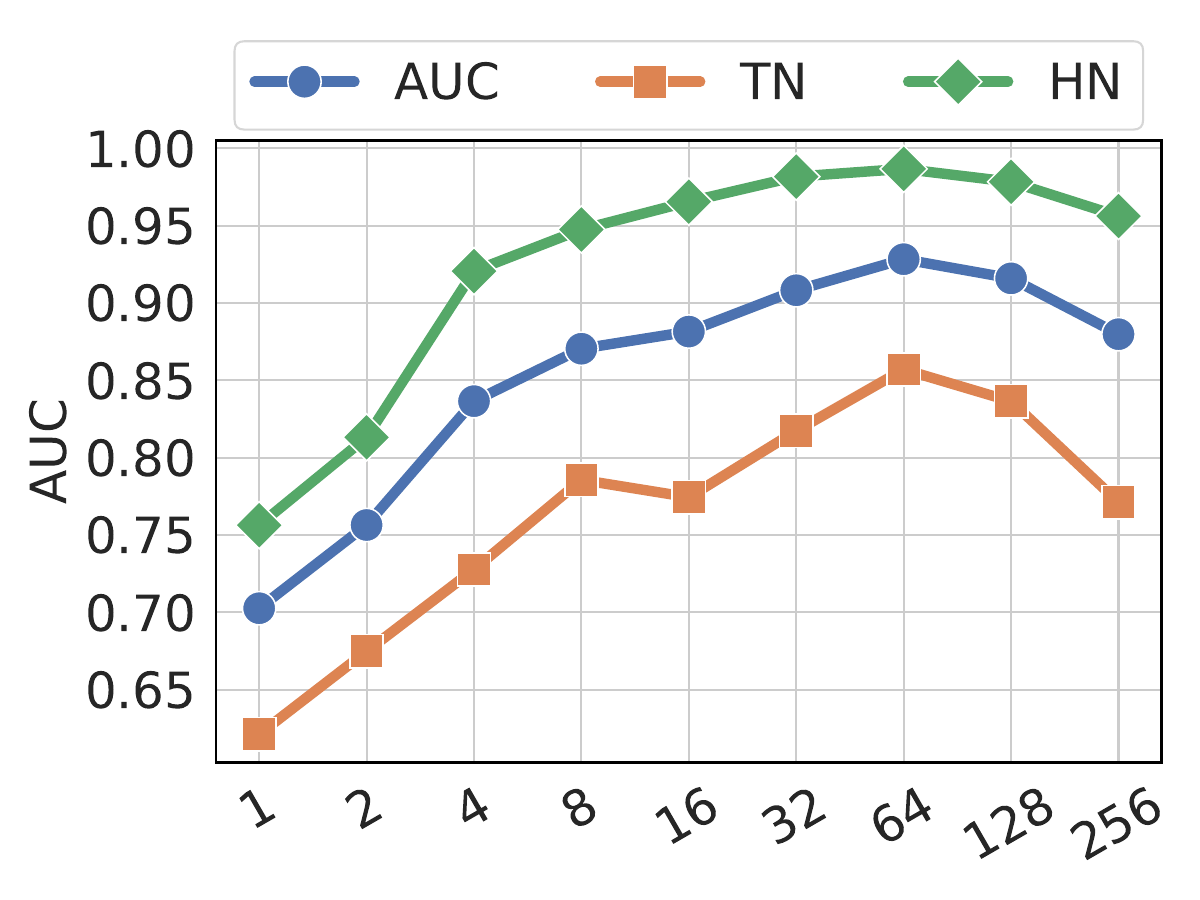}}
\subfloat[Trade-off parameter $\alpha$]{\label{fig:param3}\includegraphics[width=0.25\linewidth]{
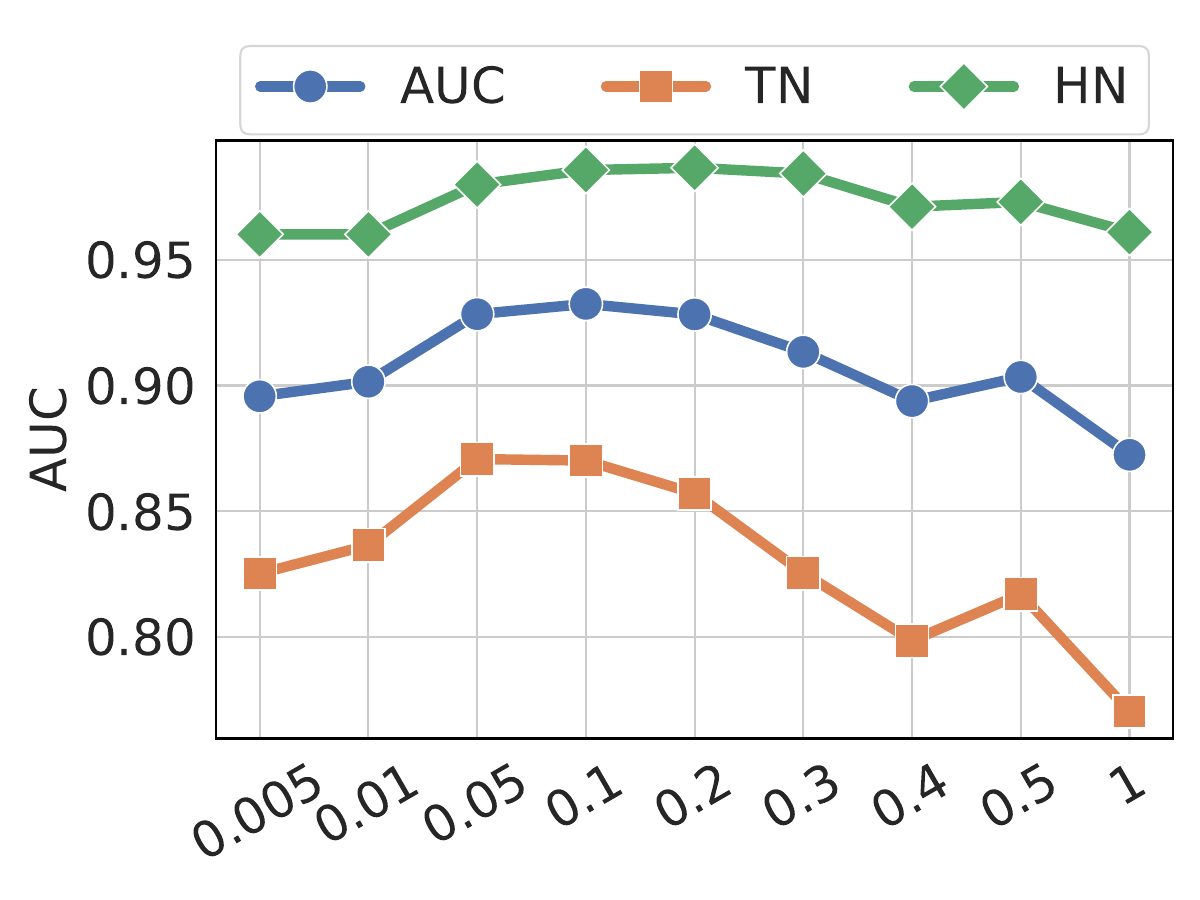}}
\subfloat[Sliding window $w$]{\label{fig:param4}\includegraphics[width=0.25\linewidth]{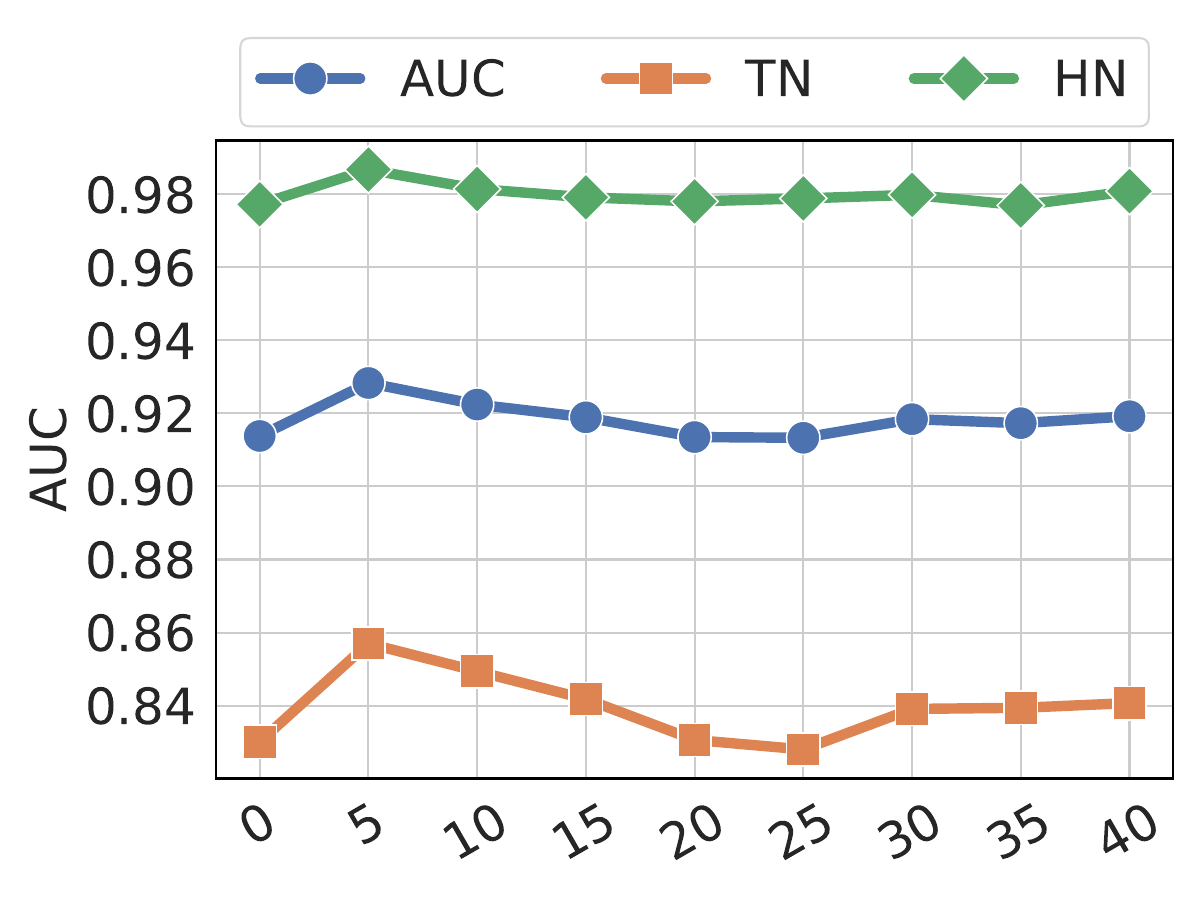}}	
\caption{The parameter study of AD-GCL with varying (a) sampling rounds $R$, (b) dimension $d$, (c)  trade-off parameter $\alpha$ and (d) sliding window $w$ on Cora dataset w.r.t. AUC respectively.}
\label{fig:param}
\end{figure*}

\subsection{Ablation Study}
To validate the validity of each component in AD-GCL, we investigate the impact of neighbor pruning (NP), anomaly-guided neighbor completion (NC), intra-view contrast loss (Intra-CL), and inter-view contrast loss (Inter-CL) in our method. W/o means removing the component during the training stage. Figure~\ref{fig:ablation} reports the results of our ablation study on six datasets. We can observe that AD-GCL consistently outperforms the other variants in terms of overall performance, as well as tail and head nodes. On the tail node, w/o NP, NC, Intra-CL and Inter-CL cause performance degradation in 2.46\%, 1.87\%, 30.62\%, and 1.89\% respectively. 
The performance is a drastic drop when w/o Intra-CL, which involves learning the latent relationships between nodes in graphs.
This observation provides evidence that intra-view contrast plays a dominant role in GCL-based anomaly detection. Meanwhile, we find an improvement in the tail performance of the NP strategy acting on the head node, further proving that aligning head nodes with forged tail nodes enhances the model's ability to learn from genuine tail nodes.
In summary, the absence of any component in AD-GCL leads to a degradation in performance, highlighting the importance of each component in the AD-GCL. \looseness=-1

\subsection{Parameter Study}
\textbf{Effect of sampling rounds $R$}\quad We investigate the effect of varying the value of $R$ in Eq.~\eqref{eq:asc}, as shown in Figure~\ref{fig:param1}. The performance is subpar when only using lower sample rounds such as 1, 2, 4, etc. As the number of sampling rounds increases, the AUC score generally increases. However, when $R$ reaches 256, further increases do not lead to significant performance changes.

\textbf{Effect of hidden layer dimension $d$}\quad We visualize the effect of the hidden layer dimension $d$ in Figure~\ref{fig:param2}. As the embedding dimension increases from 1 to 64, all three curves exhibit an overall upward trend. However, further increasing $d$, the performance begins to decline. This phenomenon can be attributed to the fact that the features in the Cora dataset are high-dimensional and sparse. When $d$ becomes too large, it may retain noise information that is not beneficial for anomaly detection tasks.

\textbf{Effect of trade-off parameter $\alpha$}\quad 
In Figure~\ref{fig:param3}, we present the evaluation results for different values of $\alpha$ in Eq.~\eqref{eq:loss}. We observe similar conclusions to the ablation experiments. Firstly, increasing $\alpha$ leads to a significant performance improvement, demonstrating the effectiveness of $\mathcal{L}_{inter}$. Additionally, from Figure~\ref{fig:ablation} we can clearly see the importance of $\mathcal{L}_{intra}$ for anomaly detection. However, when $\alpha$ becomes excessively large, $\mathcal{L}_{inter}$ dominates the overall loss, leading to a decline in performance.

\textbf{Effect of sliding window $w$}\quad We study the sliding window $w$ in the neighbor completion strategy, which incorporates anomaly scores from the previous $w$ training rounds to compute the anomaly similarity distribution. Figure~\ref{fig:param4} shows the best performance is achieved when $w=5$, and there is a significant performance drop when $w=0$, demonstrating the importance of considering anomaly scores for anomaly detection. 
However, an excessively large $w$ may introduce noise or smooth more accurate anomaly information from the current epoch, leading to performance degradation.

\section{Conclusion}
In this paper, we first investigate the limitations of anomaly detection methods based on GCL from the perspective of structural imbalance, revealing that prior works perform poor performance in detecting tail anomalies. To address the above challenges, we propose a new method AD-GCL. It filters the noisy edges of the head node through neighbor pruning and enlarges the neighborhood of the tail node through anomaly-guided neighbor completion. Finally, intra- and inter-view consistency losses of the original and enhanced graphs are introduced to enhance the representation. Extensive experiments on six benchmark datasets demonstrate that AD-GCL outperforms the competitors.

\section{Acknowledgments}
This research was partially supported by the National Key Research and Development Project of China No. 2021ZD0110700, the Key Research and Development Project in Shaanxi Province No. 2022GXLH01-03, the National Science Foundation of China No. (62250009, 62037001, 62302380, and 62476215), the China Postdoctoral Science Foundation No. 2023M742789, the Fundamental Scientific Research Funding No. (xzd012023061 and xpt012024003), and the Shaanxi Continuing Higher Education Teaching Reform Research Project No. 21XJZ014. Co-author Chen Chen consulted on this project on unpaid weekends for personal interests, and appreciated collaborators and family for their understanding.

\bibliography{aaai25}

\newpage
\begin{center}
    \huge \textbf{Appendix} \label{sec:appendix}
\end{center}
\appendix
\renewcommand\thefigure{A\arabic{figure}}
\renewcommand\theequation{A\arabic{equation}}
\frenchspacing
\maketitle

\section{Complexity Analysis}\label{sec:complexity_analysis}
The time complexity of existing GCL-based anomaly detection methods is primarily determined by the GNN encoder and the contrastive loss, with time complexities denoted as $O\left(\left|\mathcal{V}\right|d^{2}+\left| \mathcal{E}\right|d\right)$ and $O\left ( \left| \mathcal{V}\right|^{2}d \right)$, respectively, where $\mathcal{V}$ is the set of nodes, $\mathcal{E}$ is the sets of  edges. AD-GCL introduces additional modules, namely the Neighbor Pruning and Neighbor Completion, which incur additional time complexities of $O\left(\left|\mathcal{V}\right|^{2}d\right)$ and $O\left (\left|\mathcal{V} \right|^{2}w \right ) $, respectively, where sliding window $w\ll d$. The overall time complexity of AD-GCL is $O\left(\left|\mathcal{V}\right|d^{2}+\left|\mathcal{E}\right|d+\left| \mathcal{V}\right|^{2}d\right)$, which is approximately equal to the time complexity of existing  GCL-base anomaly detection models while increasing detection effectiveness. \looseness=-1

\section{Experimental Details}
\subsection{Datasets details}\label{sec:appendix_datasets}
\textbf{Citation Networks}~\cite{liu2021anomaly,jin2021anemone}. Cora, Citeseer, and Pubmed are publicly available datasets comprising scientific publications. In these networks, nodes represent individual papers, while edges depict citation relationships between papers. The attribute vector of each node is the bag-of-word representation, with the dimension determined by the size of the dictionary.

\textbf{Bitcoin Trading Networks}~\cite{kumar2016edge, kumar2018rev2}. Bitcoin is a cryptocurrency used for anonymous transactions on the web. Bitcoinotc, BITotc, and BITalpha are three datasets from two bitcoin trading platforms, OTC and alpha~\footnote{\url{https://www.bitcoin-otc.com/} and \url{https://btc-alpha.com/}}, where nodes represent Bitcoin users, while edges describe the transaction relationship between Bitcoin users.

\textbf{Social Network}~\cite{liu2022bond,tang2024gadbench}. Reddit is a user-subreddit graph extracted from a social media platform, Reddit~\footnote{\url{https://www.reddit.com/}}. This public dataset consists of one month of user posts on subreddits~\footnote{\url{http://files.pushshift.io/reddit/}}. The dataset consists of 1,000 most active subreddits and the 10,000 most active users. The dataset contains 168,016 interactions. Each user has a binary label indicating whether it has been banned by the platform. Textual posts are converted into feature vectors representing LIWC categories~\cite{pennebaker2001linguistic}, and the features of users and subreddits are computed as the feature summation of the posts they have, respectively.

\textbf{Worker Relationship Network}~\cite{platonov2023critical,tang2024gadbench}. The Tolokers dataset originates from the Toloka crowdsourcing platform. Each node represents a toloker (worker) who has participated in at least one of the 13 selected projects. An edge between two tolokers if they have worked on the same task. The objective is to predict which tolokers have been banned in one of the projects. Node features include the worker's profile information and task performance statistics.

\begin{figure*}
\centering
\subfloat[Cora]{\label{fig:dis1}\includegraphics[width=0.25\linewidth]{
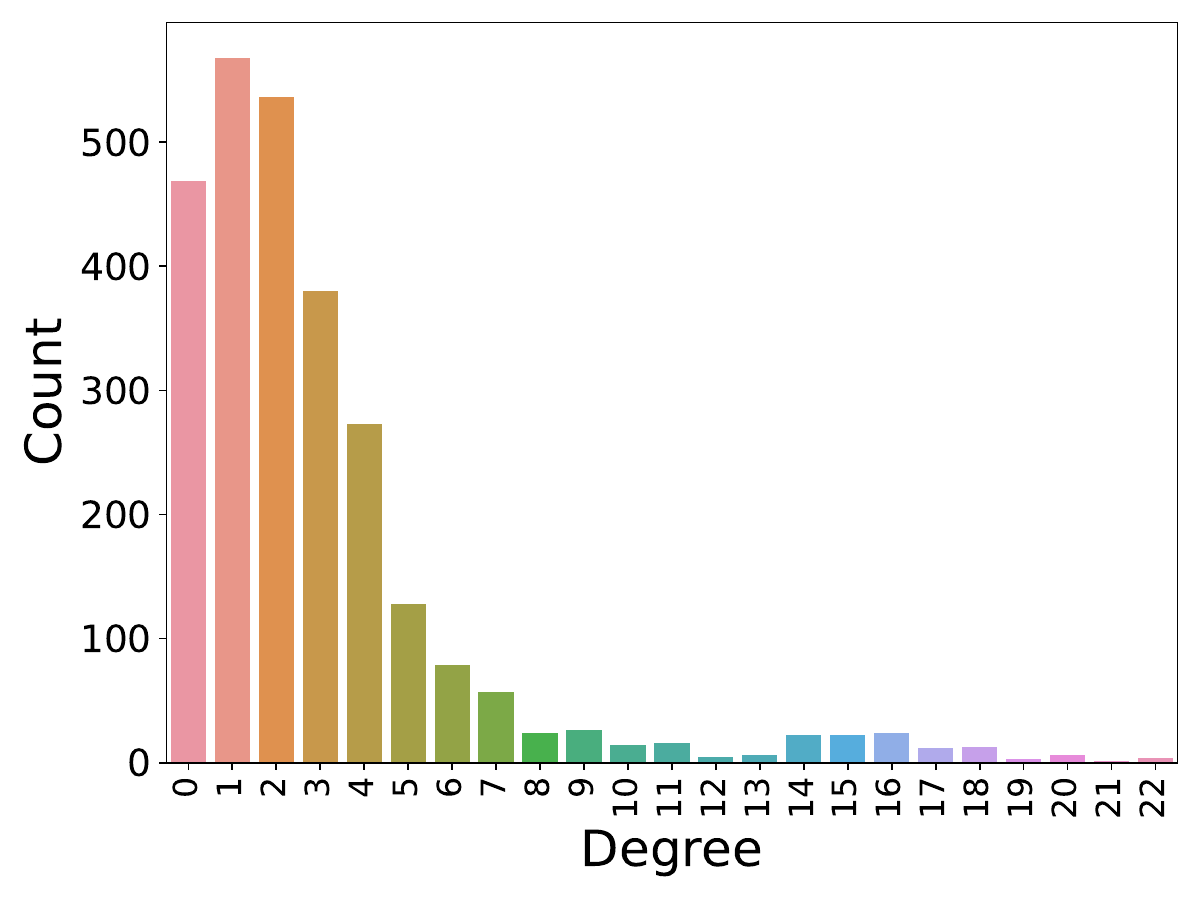}}
\subfloat[Citeseer]{\label{fig:dis2}\includegraphics[width=0.25\linewidth]{
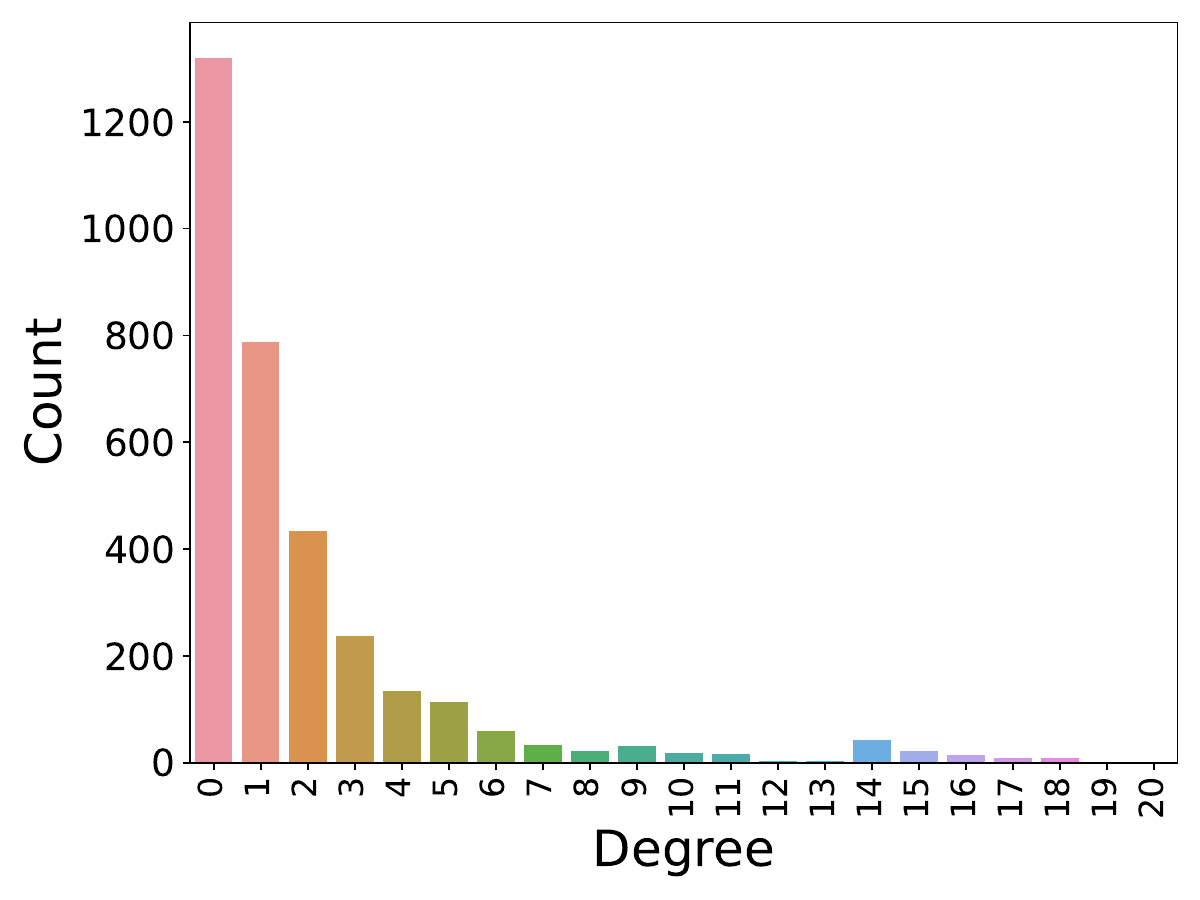}}
\subfloat[Pubmed]{\label{fig:dis3}\includegraphics[width=0.25\linewidth]{
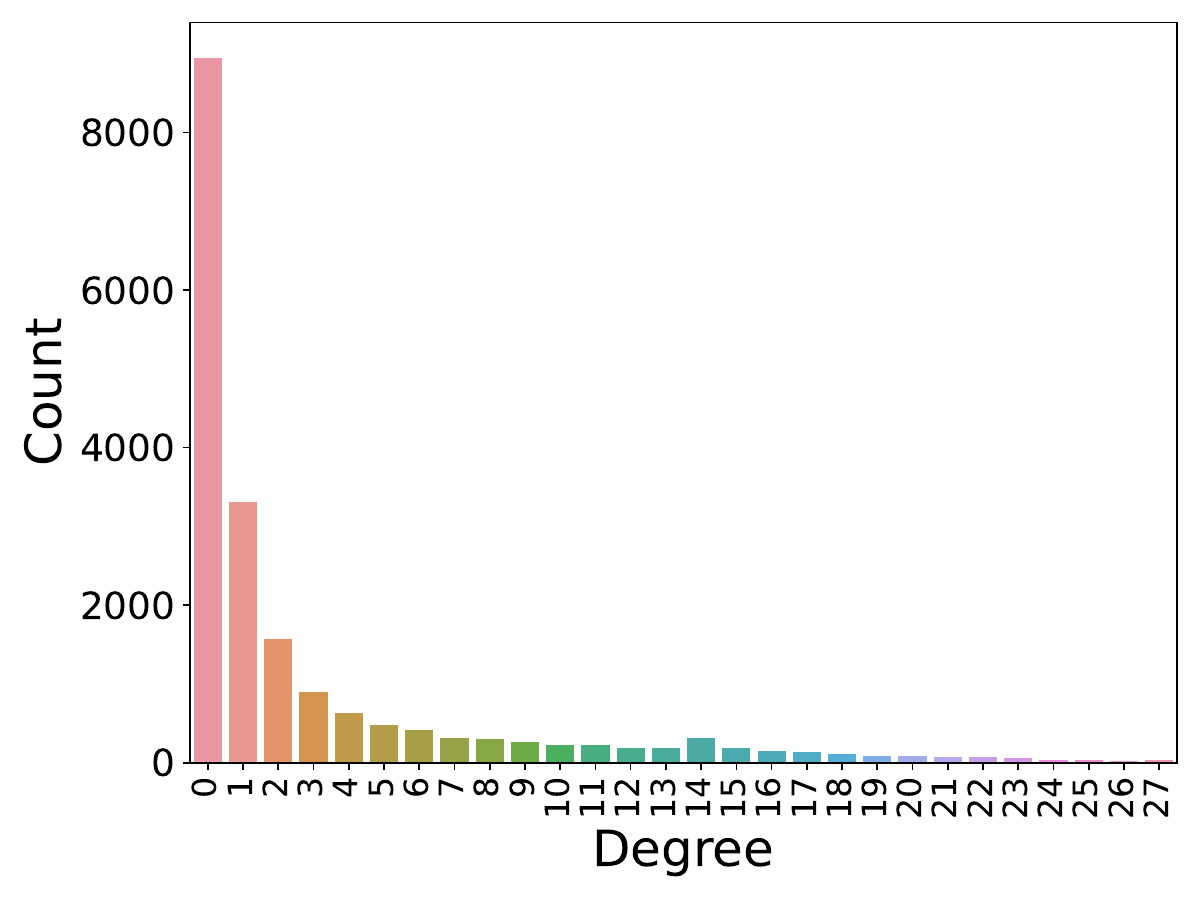}}
\subfloat[Bitcoinotc]{\label{fig:dis4}\includegraphics[width=0.25\linewidth]{
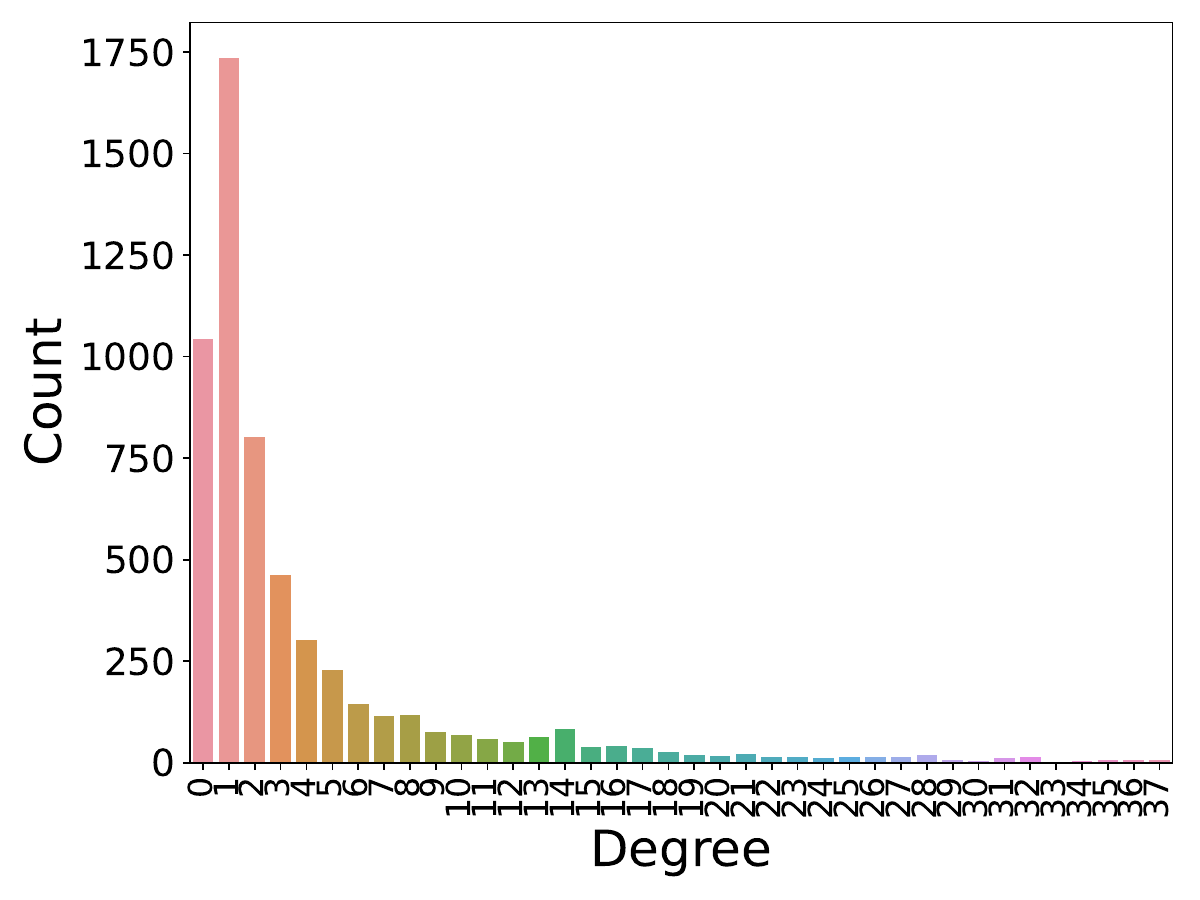}}\\
\subfloat[BITotc]{\label{fig:dis5}\includegraphics[width=0.25\linewidth]{
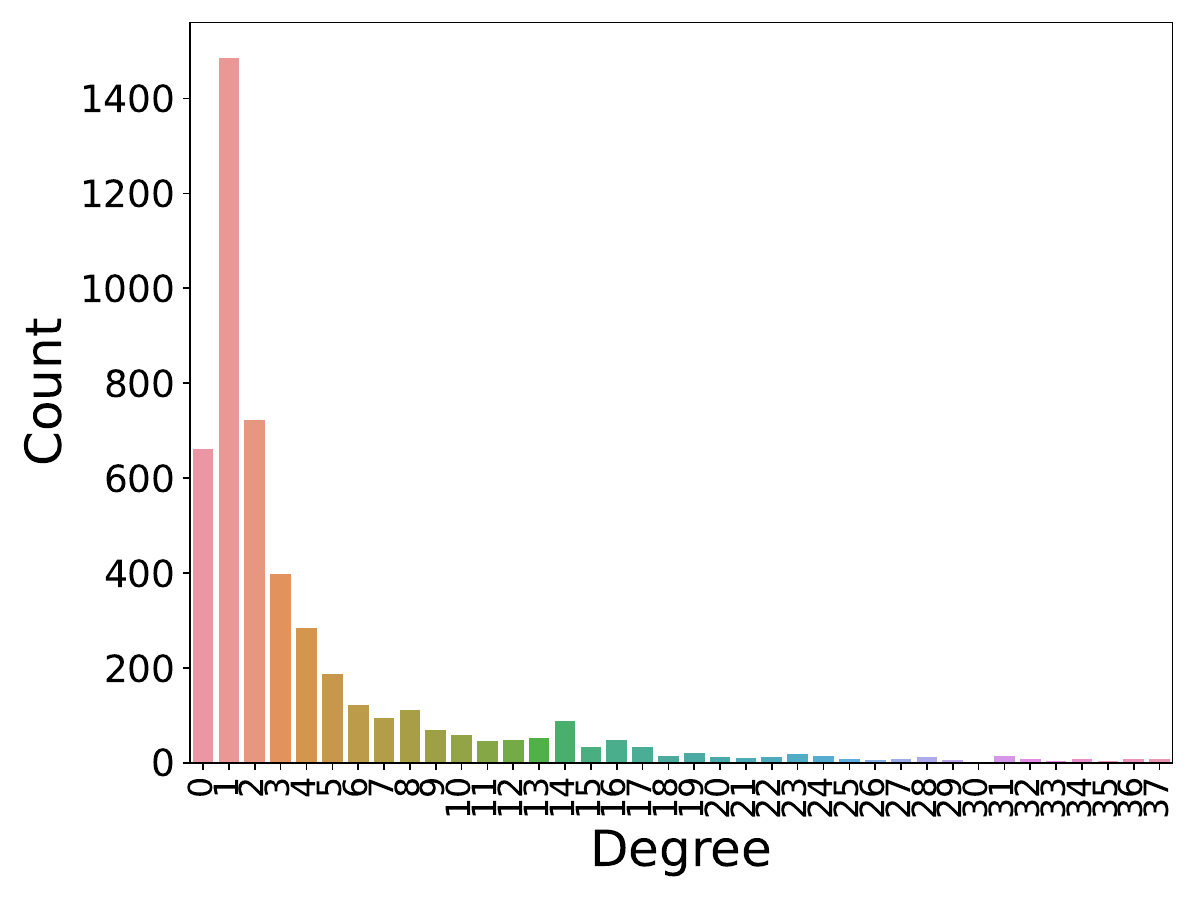}}
\subfloat[BITalpha]{\label{fig:dis6}\includegraphics[width=0.25\linewidth]{
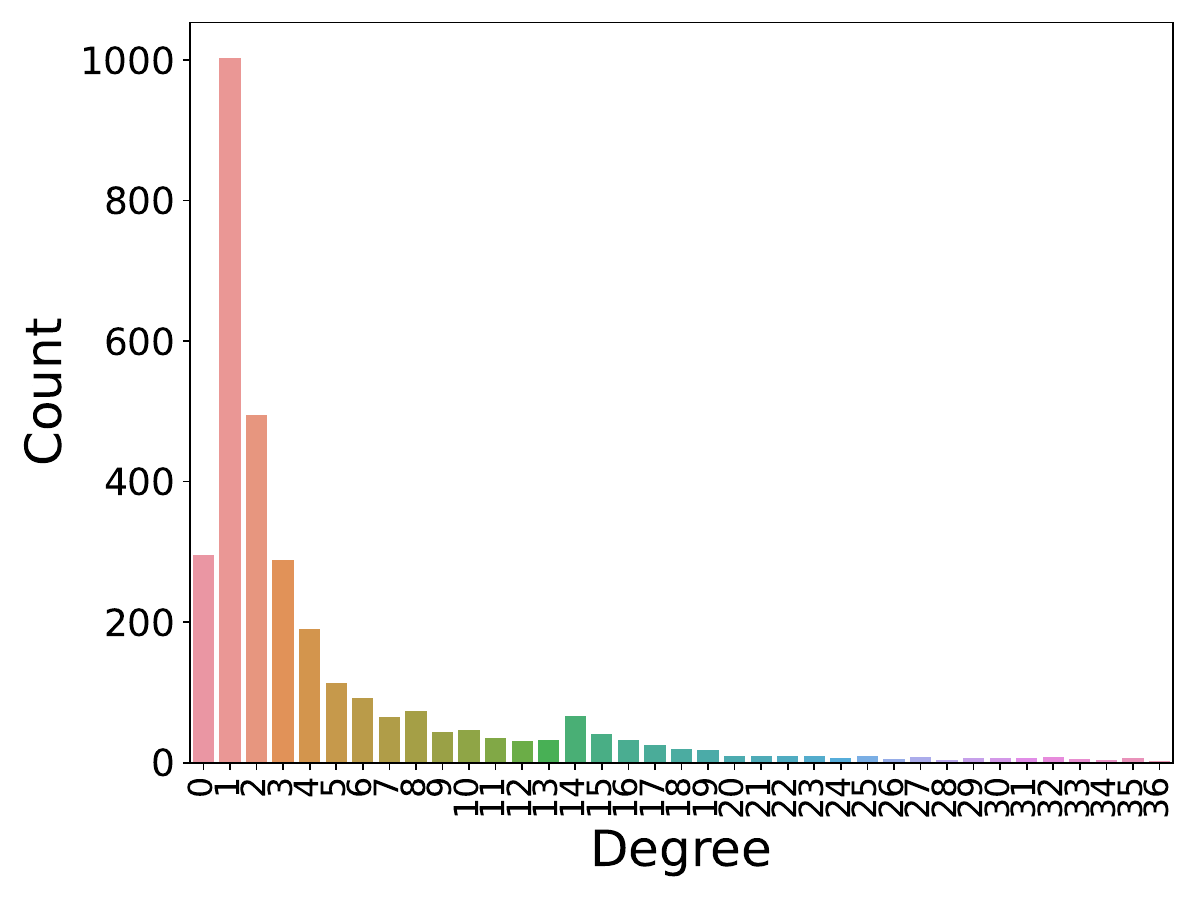}}
\subfloat[Reddit]{\label{fig:dis7}\includegraphics[width=0.25\linewidth]{
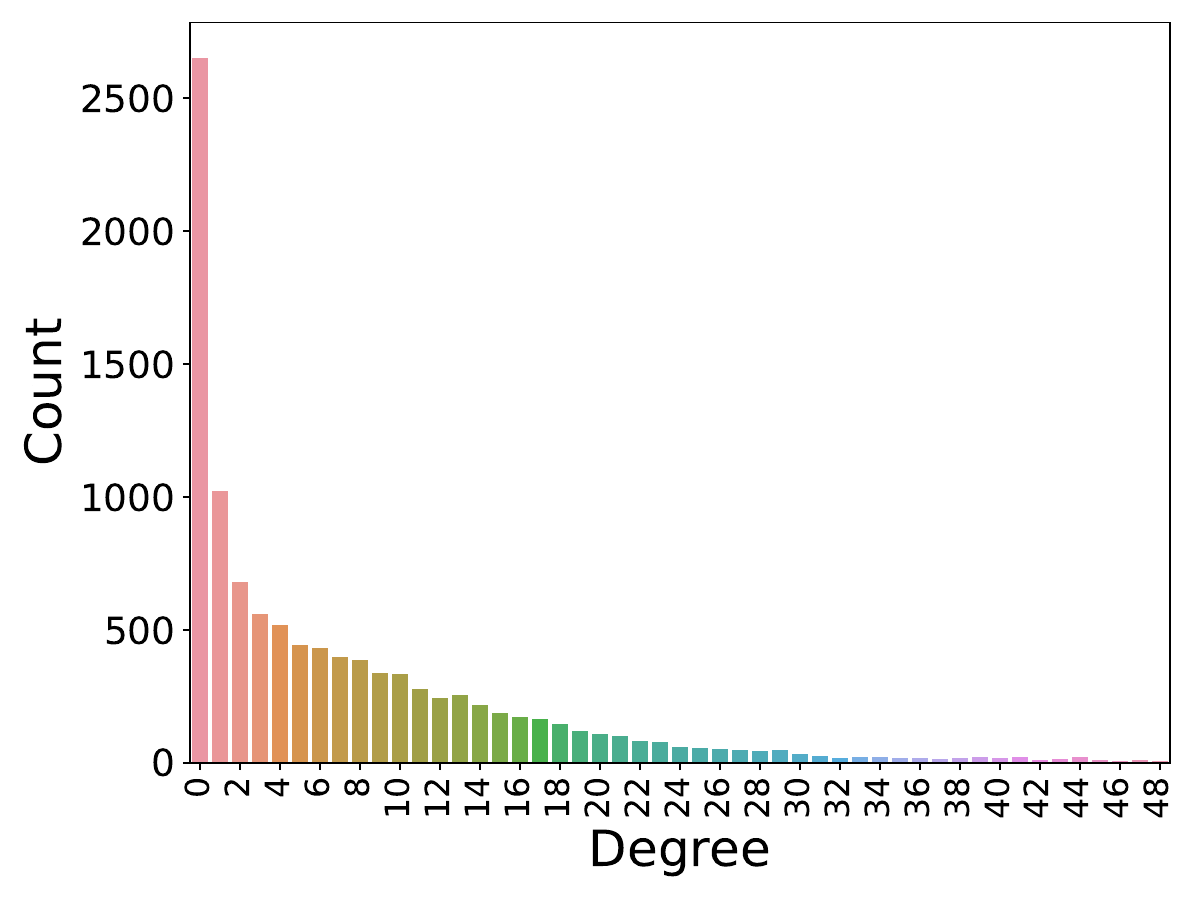}}
\subfloat[Tolokers]{\label{fig:dis8}\includegraphics[width=0.25\linewidth]{
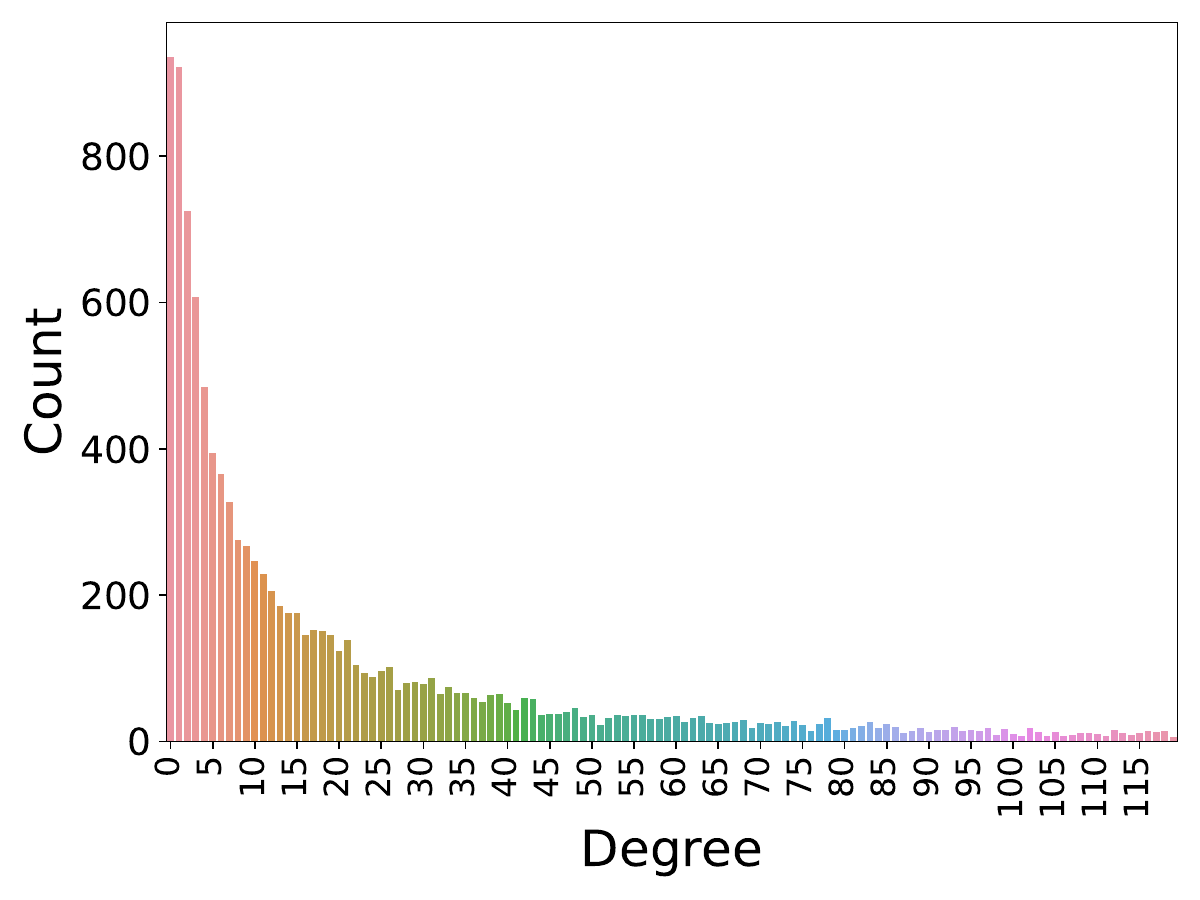}}\\
\caption{Degree distributions of real-world graphs.}
\label{fig:dis}    
\end{figure*}

\begin{table*}[t!]
\renewcommand\arraystretch{1.2} 
\centering
  \caption{Statistics of the Datasets.}
  \label{tab:SD}
\resizebox{0.75\textwidth}{!}{
  \begin{tabular}{ccccccc}
    \toprule
    \# Dataset & \# Nodes & \# Edges & \# Attributes & \# Anomalies & \# Tail anomalies & \# Head anomalies \\
    \midrule
    Cora & 2,708 & 5,429 & 1,433 & 5.5\% & 2.3\% & 3.2\% \\
    Citeseer & 3,327 & 4,732 & 3,703 & 4.5\% & 2.2\% & 2.3\% \\
    Pubmed & 19,717 & 44,338 & 500 & 3.0\% & 1.2\% & 1.8\% \\
    Bitcoinotc & 5,881 & 35,592 & 128 & 5.1\% & 1.9\% & 3.2\% \\
    BITotc & 4,863 & 28,473 & 128 & 6.2\% & 2.3\% & 3.8\% \\
    BITalpha & 3,219 & 19,364 & 128 & 9.3\% & 3.4\% & 5.9\% \\
    Reddit & 10,984 & 168,016 & 64 & 3.3\% & 1.8\% & 1.5\% \\
    Tolokers & 11,758 & 519,000 & 10 & 21.8\% & 17.8\% & 4.0\% \\
    \bottomrule
  \end{tabular}}
\end{table*}

We visualize the degree distributions across eight real-world datasets, as depicted in Figure~\ref{fig:dis}. Through this visualization, we can observe that real-world graphs adhere to a power-law distribution, where a significant portion of nodes exhibit low degrees. The statistics of these datasets are summarized in Table~\ref{tab:SD}. 

Reddit~\cite{liu2022bond,tang2024gadbench} and Tolokers~\cite{platonov2023critical,tang2024gadbench} are real-world datasets containing anomalies, given the absence of anomalies in the other six datasets by default. To assess efficacy of our method in detecting diverse anomaly types, we adhere to previous methodologies~\cite{song2007conditional,ding2019deep,liu2021anomaly} by manually introducing an equal number of attributive and structural anomalous nodes into these six datasets. We provide a detailed supplementary description of how to inject head and tail anomalies into the dataset. Specifically, for structural anomalies, in a small clique, a small set of nodes are much more closely linked to each other than average, aligning with typical structural abnormalities observed in real-world networks~\cite{skillicorn2007detecting}. To simulate such anomalies, we commence by defining the clique size $p$ and the number of cliques $q$. When generating a clique,  $p$ nodes are randomly selected from the node set $\mathcal{V}$ and fully connected, thus marking all selected nodes $p$ as structural anomaly nodes. This process is iterated $q$ times to generate $q$ cliques, resulting in a total injection of  $p \times q$ structural anomalies. For feature anomalies, inconsistencies between node features and their neighboring contexts represent another prevalent anomaly observed in real-world scenarios. Following the pattern introduced by~\cite{song2007conditional}, feature anomalies are created by perturbing node attributes. When generating a single feature anomaly, we first select a target node $v_i$ and then sample a set of $k$ nodes as candidates. Subsequently, from the candidate set, we choose the node $v_j$ with the maximum Euclidean distance from the feature of the target node $v_i$ , and replace $v_i$'s feature with that of $v_j$. Here, we set $k=50$ to ensure a sufficiently large perturbation magnitude. To ensure an equal balance in the quantity of both anomaly types, we set the number of feature anomalies to $p \times q$, implying that the above operation is repeated $p \times q$ times to generate all feature anomalies. As shown in Table~\ref{tab:SD}, both strategies introduce a sufficient number of head and tail anomaly nodes (classified as head or tail based solely on their degree).

\subsection{Implementation Details} \label{sec:appendix_param}
We adopt GCN~\cite{kipf2016semi} as the encoder. Node features are mapped to 64 dimensions in hidden space. Additionally, we calculate anomaly scores for $R=256$ rounds. We set sliding window $w=5$, temperature parameter $\tau$ is 0.07 and trade-off parameter $\alpha=0.2$. Adam optimizer~\cite{kingma2014adam} is used in the training model. The learning rates are as follows: Cora (5e-3), Citeseer (3e-3), Pubmed (4e-3), Bitcoinotc (4e-4), BITotc (5e-4), BITalpha (5e-3), Reddit (5e-4), and Tolokers (4e-2). We specify model training for 200 epochs on the Cora and Citeseer datasets, and model training for 300 epochs on Reddit and Tolokers, while for the Pubmed, Bitcoinotc, BITotc, and BITalpha datasets, the training is conducted for 100 epochs. 
The selection of the predefined threshold $K$ follows previous work~\cite{liu2020towards,liu2021tail} and the Pareto principle~\cite{sanders1987pareto}, where $K$ is set to 6 for the Cora, Citeseer, Pubmed, Bitcoinotc, BITotc, and BITalpha datasets, and $K$ is set to 9 and 90 for the more dense graph Reddit and Tolokers datasets, respectively. We report the AUC and the standard deviation with all the experiments run 5 times using different random seeds.

\begin{table*}[ht]
\centering
\renewcommand\arraystretch{1.3}
\setlength\tabcolsep{3pt}
\caption{Performance comparison for all nodes (top), tail nodes (middle), and head nodes (bottom) on unsupervised anomaly detection setting. We report both AUPRC and AP. Bold indicates the optimal and underline indicates the suboptimal.}
\resizebox{1\textwidth}{!}{
\begin{tabular}{ccccccccccccccc}
\toprule
\multirow{2}{*}{All} & \multicolumn{2}{c}{Pubmed} & \multicolumn{2}{c}{Bitcoinotc} & \multicolumn{2}{c}{BITotc} & \multicolumn{2}{c}{BITalpha} & \multicolumn{2}{c}{Reddit} & \multicolumn{2}{c}{Tolokers} \\
\cline{2-13}
 & AUPRC & AP & AUPRC & AP & AUPRC & AP & AUPRC & AP & AUPRC & AP & AUPRC & AP \\
\hline
AEGIS
& 5.70$_{\pm 0.50}$  & 5.81$_{\pm 0.12}$ 
& 5.35$_{\pm 0.35}$  & 5.39$_{\pm 0.38}$ 
& 6.55$_{\pm 0.47}$  & 6.62$_{\pm 0.86}$ 
& 9.60$_{\pm 0.81}$  & 9.67$_{\pm 0.61}$ 
& 3.44$_{\pm 0.89}$  & 3.47$_{\pm 0.17}$ 
& $\underline{28.46_{\pm 0.65}}$  & $\underline{28.49_{\pm 0.28}}$  \\

GAAN
& 3.14$_{\pm 0.81}$  & 3.16$_{\pm 0.06}$ 
& 6.59$_{\pm 0.74}$  & 6.68$_{\pm 0.20}$ 
& 7.32$_{\pm 0.18}$  & 7.44$_{\pm 0.20}$ 
& 11.34$_{\pm 0.14}$  & 11.44$_{\pm 0.44}$ 
& 3.37$_{\pm 0.53}$  & 3.39$_{\pm 0.23}$ 
& 25.59$_{\pm 0.32}$  & 25.62$_{\pm 0.63}$  \\

CoLA
& 30.00$_{\pm 0.44}$  & 30.17$_{\pm 0.32}$ 
& 20.50$_{\pm 0.35}$  & 20.71$_{\pm 0.31}$ 
& 23.40$_{\pm 0.54}$  & 23.60$_{\pm 0.50}$ 
& 21.55$_{\pm 0.68}$  & 21.69$_{\pm 0.09}$ 
& $\underline{4.24_{\pm 0.70}}$  & $\underline{4.29_{\pm 0.23}}$ 
& 26.86$_{\pm 0.64}$  & 26.89$_{\pm 0.41}$  \\

ANEMONE
& $\underline{44.25_{\pm 0.78}}$  & $\underline{44.41_{\pm 0.65}}$ 
& 25.84$_{\pm 0.50}$  & 26.14$_{\pm 0.02}$ 
& $\underline{28.71_{\pm 0.13}}$  & $\underline{28.98_{\pm 0.26}}$ 
& 22.07$_{\pm 0.39}$  & 22.24$_{\pm 0.80}$ 
& 3.95$_{\pm 0.52}$  & 3.99$_{\pm 0.09}$ 
& 26.12$_{\pm 0.16}$  & 26.16$_{\pm 0.64}$  \\

AdONE
& 14.04$_{\pm 0.33}$  & 14.12$_{\pm 0.81}$ 
& 16.10$_{\pm 0.76}$  & 16.28$_{\pm 0.41}$ 
& 20.04$_{\pm 0.05}$  & 20.15$_{\pm 0.83}$ 
& $\underline{28.85_{\pm 0.14}}$  & $\underline{29.02_{\pm 0.34}}$ 
& 3.74$_{\pm 0.14}$  & 3.76$_{\pm 0.42}$ 
& 22.70$_{\pm 0.78}$  & 22.73$_{\pm 0.64}$  \\

GRADATE
& 39.41$_{\pm 0.61}$  & 39.47$_{\pm 0.32}$ 
& $\underline{27.44_{\pm 0.77}}$  & $\underline{27.69_{\pm 0.47}}$ 
& 27.56$_{\pm 0.62}$  & 27.86$_{\pm 0.88}$ 
& 22.76$_{\pm 0.58}$  & 22.97$_{\pm 0.45}$ 
& 3.07$_{\pm 0.19}$  & 3.14$_{\pm 0.26}$ 
& 21.44$_{\pm 0.87}$  & 21.47$_{\pm 0.32}$  \\

GAD-NR
& 8.87$_{\pm 0.05}$  & 8.91$_{\pm 0.45}$ 
& 11.54$_{\pm 0.89}$  & 11.70$_{\pm 0.43}$ 
& 14.24$_{\pm 0.44}$  & 15.51$_{\pm 0.19}$ 
& 21.54$_{\pm 0.65}$  & 21.68$_{\pm 0.83}$ 
& 3.66$_{\pm 0.20}$  & 3.68$_{\pm 0.82}$ 
& $\mathbf{30.20}_{\pm 0.85}$  & $\mathbf{30.21}_{\pm 0.44}$  \\
\hline

AD-GCL
& $\mathbf{58.07}_{\pm 0.10}$  & $\mathbf{58.13}_{\pm 0.08}$ 
& $\mathbf{28.94}_{\pm 0.35}$  & $\mathbf{29.20}_{\pm 0.03}$ 
& $\mathbf{29.32}_{\pm 0.44}$  & $\mathbf{29.59}_{\pm 0.86}$ 
& $\mathbf{35.86}_{\pm 0.38}$  & $\mathbf{36.11}_{\pm 0.60}$ 
& $\mathbf{5.66}_{\pm 0.31}$  & $\mathbf{5.76}_{\pm 0.71}$ 
& 27.02$_{\pm 0.38}$  & 27.05$_{\pm 0.07}$  \\

\bottomrule
\end{tabular}}

\resizebox{1\textwidth}{!}{
\begin{tabular}{ccccccccccccccc}
\toprule
\multirow{2}{*}{Tail} & \multicolumn{2}{c}{Pubmed} & \multicolumn{2}{c}{Bitcoinotc} & \multicolumn{2}{c}{BITotc} & \multicolumn{2}{c}{BITalpha} & \multicolumn{2}{c}{Reddit} & \multicolumn{2}{c}{Tolokers} \\
\cline{2-13}
 & AUPRC & AP & AUPRC & AP & AUPRC & AP & AUPRC & AP & AUPRC & AP & AUPRC & AP \\
\hline
AEGIS
& 5.45$_{\pm 0.72}$  & 5.68$_{\pm 0.62}$ 
& 3.03$_{\pm 0.06}$  & 3.08$_{\pm 0.72}$ 
& 3.94$_{\pm 0.47}$  & 4.09$_{\pm 0.08}$ 
& 5.37$_{\pm 0.51}$  & 5.47$_{\pm 0.63}$ 
& 2.65$_{\pm 0.26}$  & 2.69$_{\pm 0.82}$ 
& 23.97$_{\pm 0.65}$  & 24.14$_{\pm 0.04}$  \\

GAAN
& 1.49$_{\pm 0.87}$  & 1.50$_{\pm 0.57}$ 
& 2.62$_{\pm 0.67}$  & 2.68$_{\pm 0.07}$ 
& 3.00$_{\pm 0.10}$  & 3.06$_{\pm 0.28}$ 
& 4.60$_{\pm 0.87}$  & 4.70$_{\pm 0.44}$ 
& 2.59$_{\pm 0.71}$  & 2.63$_{\pm 0.87}$ 
& 16.72$_{\pm 0.15}$  & 16.83$_{\pm 0.59}$  \\

CoLA
& 18.67$_{\pm 0.82}$  & 19.00$_{\pm 0.15}$ 
& 3.66$_{\pm 0.86}$  & 3.77$_{\pm 0.17}$ 
& 5.56$_{\pm 0.68}$  & 5.80$_{\pm 0.43}$ 
& $\mathbf{8.64}_{\pm 0.28}$  & $\mathbf{8.90}_{\pm 0.49}$ 
& 3.44$_{\pm 0.69}$  & 3.52$_{\pm 0.68}$ 
& $\underline{25.16_{\pm 0.77}}$  & $\underline{25.30_{\pm 0.71}}$  \\

ANEMONE
& $\underline{33.21_{\pm 0.66}}$  & $\underline{33.51_{\pm 0.60}}$ 
& $\underline{3.76_{\pm 0.88}}$  & $\underline{3.84_{\pm 0.42}}$ 
& $\underline{5.95_{\pm 0.63}}$  & $\underline{6.23_{\pm 0.07}}$ 
& 7.28$_{\pm 0.53}$  & 7.49$_{\pm 0.33}$ 
& $\underline{3.64_{\pm 0.49}}$  & $\underline{3.72_{\pm 0.72}}$ 
& 23.56$_{\pm 0.47}$  & 23.75$_{\pm 0.25}$  \\

AdONE
& 13.17$_{\pm 0.28}$  & 13.41$_{\pm 0.23}$ 
& 3.59$_{\pm 0.62}$  & 3.67$_{\pm 0.49}$ 
& 5.56$_{\pm 0.32}$  & 5.67$_{\pm 0.88}$ 
& 5.43$_{\pm 0.06}$  & 5.57$_{\pm 0.55}$ 
& 3.04$_{\pm 0.62}$  & 3.08$_{\pm 0.49}$ 
& 15.02$_{\pm 0.09}$  & 15.09$_{\pm 0.80}$  \\

GRADATE
& 17.29$_{\pm 0.30}$  & 17.45$_{\pm 0.21}$ 
& 2.77$_{\pm 0.88}$  & 2.84$_{\pm 0.81}$ 
& 4.21$_{\pm 0.78}$  & 4.33$_{\pm 0.77}$ 
& 4.96$_{\pm 0.65}$  & 5.17$_{\pm 0.09}$ 
& 2.42$_{\pm 0.78}$  & 2.62$_{\pm 0.89}$ 
& 15.65$_{\pm 0.03}$  & 15.82$_{\pm 0.12}$  \\

GAD-NR
& 1.34$_{\pm 0.76}$  & 1.36$_{\pm 0.68}$ 
& 2.49$_{\pm 0.17}$  & 2.56$_{\pm 0.85}$ 
& 2.84$_{\pm 0.51}$  & 3.01$_{\pm 0.55}$ 
& 5.56$_{\pm 0.15}$  & 5.64$_{\pm 0.15}$ 
& 3.22$_{\pm 0.63}$  & 3.32$_{\pm 0.17}$ 
& 15.89$_{\pm 0.02}$  & 16.23$_{\pm 0.35}$  \\
\hline

AD-GCL
& $\mathbf{49.53}_{\pm 0.79}$  & $\mathbf{49.69}_{\pm 0.81}$ 
& $\mathbf{5.04}_{\pm 0.81}$  & $\mathbf{5.22}_{\pm 0.66}$ 
& $\mathbf{6.18}_{\pm 0.52}$  & $\mathbf{6.39}_{\pm 0.54}$ 
& $\underline{8.09_{\pm 0.21}}$  & $\underline{8.42_{\pm 0.33}}$ 
& $\mathbf{4.27}_{\pm 0.84}$  & $\mathbf{4.38}_{\pm 0.20}$ 
& $\mathbf{26.10}_{\pm 0.69}$  & $\mathbf{26.23}_{\pm 0.06}$  \\
\bottomrule
\end{tabular}}

\resizebox{1\textwidth}{!}{
\begin{tabular}{ccccccccccccccc}
\toprule
\multirow{2}{*}{Head} & \multicolumn{2}{c}{Pubmed} & \multicolumn{2}{c}{Bitcoinotc} & \multicolumn{2}{c}{BITotc} & \multicolumn{2}{c}{BITalpha} & \multicolumn{2}{c}{Reddit} & \multicolumn{2}{c}{Tolokers} \\
\cline{2-13}
 & AUPRC & AP & AUPRC & AP & AUPRC & AP & AUPRC & AP & AUPRC & AP & AUPRC & AP \\
\hline
AEGIS
& 9.71$_{\pm 0.59}$  & 9.83$_{\pm 0.31}$ 
& 11.80$_{\pm 0.10}$  & 11.99$_{\pm 0.74}$ 
& 15.20$_{\pm 0.38}$  & 15.41$_{\pm 0.46}$ 
& 19.64$_{\pm 0.80}$  & 19.84$_{\pm 0.21}$ 
& 4.15$_{\pm 0.27}$  & 4.22$_{\pm 0.79}$ 
& $\underline{29.65_{\pm 0.62}}$  & $\underline{29.69_{\pm 0.25}}$  \\

GAAN
& 8.71$_{\pm 0.30}$  & 8.78$_{\pm 0.08}$ 
& 12.60$_{\pm 0.05}$  & 12.77$_{\pm 0.58}$ 
& 16.01$_{\pm 0.21}$  & 16.39$_{\pm 0.75}$ 
& 22.34$_{\pm 0.08}$  & 22.59$_{\pm 0.17}$ 
& 3.98$_{\pm 0.19}$  & 4.02$_{\pm 0.78}$ 
& 26.70$_{\pm 0.78}$  & 26.73$_{\pm 0.71}$  \\

CoLA
& 79.90$_{\pm 0.69}$  & 79.94$_{\pm 0.46}$ 
& 36.91$_{\pm 0.77}$  & 37.30$_{\pm 0.46}$ 
& 39.23$_{\pm 0.44}$  & 39.58$_{\pm 0.87}$ 
& 30.04$_{\pm 0.84}$  & 30.31$_{\pm 0.19}$ 
& $\underline{5.11_{\pm 0.46}}$  & $\underline{5.30_{\pm 0.18}}$ 
& 27.06$_{\pm 0.31}$  & 27.09$_{\pm 0.16}$  \\

ANEMONE
& $\underline{87.68_{\pm 0.57}}$  & $\underline{87.70_{\pm 0.30}}$ 
& 45.08$_{\pm 0.35}$  & 45.43$_{\pm 0.78}$ 
& $\underline{46.17_{\pm 0.84}}$  & $\underline{46.60_{\pm 0.67}}$ 
& 34.63$_{\pm 0.28}$  & 34.93$_{\pm 0.84}$ 
& 4.38$_{\pm 0.42}$  & 4.47$_{\pm 0.04}$ 
& 26.41$_{\pm 0.07}$  & 26.45$_{\pm 0.84}$  \\

AdONE
& 14.85$_{\pm 0.39}$  & 14.97$_{\pm 0.76}$ 
& 20.69$_{\pm 0.36}$  & 20.96$_{\pm 0.40}$ 
& 25.26$_{\pm 0.64}$  & 25.43$_{\pm 0.84}$ 
& $\underline{40.19_{\pm 0.48}}$  & $\underline{40.46_{\pm 0.04}}$ 
& 4.16$_{\pm 0.31}$  & 4.20$_{\pm 0.14}$ 
& 23.91$_{\pm 0.15}$  & 23.94$_{\pm 0.84}$  \\

GRADATE
& 86.84$_{\pm 0.30}$  & 86.85$_{\pm 0.44}$ 
& $\underline{45.65_{\pm 0.55}}$  & $\underline{46.04_{\pm 0.05}}$ 
& 43.39$_{\pm 0.37}$  & 43.86$_{\pm 0.05}$ 
& 37.73$_{\pm 0.02}$  & 38.09$_{\pm 0.31}$ 
& 4.51$_{\pm 0.64}$  & 4.58$_{\pm 0.10}$ 
& 21.85$_{\pm 0.43}$  & 21.89$_{\pm 0.07}$  \\

GAD-NR
& 12.07$_{\pm 0.28}$  & 12.14$_{\pm 0.04}$ 
& 14.40$_{\pm 0.09}$  & 14.64$_{\pm 0.22}$ 
& 17.98$_{\pm 0.54}$  & 20.12$_{\pm 0.61}$ 
& 26.78$_{\pm 0.67}$  & 26.99$_{\pm 0.06}$ 
& 4.31$_{\pm 0.86}$  & 4.35$_{\pm 0.72}$ 
& $\mathbf{31.62}_{\pm 0.54}$  & $\mathbf{31.63}_{\pm 0.76}$  \\
\hline
AD-GCL
& $\mathbf{89.08}_{\pm 0.44}$  & $\mathbf{89.10}_{\pm 0.30}$ 
& $\mathbf{47.42}_{\pm 0.60}$  & $\mathbf{47.84}_{\pm 0.16}$ 
& $\mathbf{48.07}_{\pm 0.72}$  & $\mathbf{48.49}_{\pm 0.51}$ 
& $\mathbf{47.88}_{\pm 0.43}$  & $\mathbf{48.26}_{\pm 0.46}$ 
& $\mathbf{6.46}_{\pm 0.28}$  & $\mathbf{6.62}_{\pm 0.06}$ 
& 27.14$_{\pm 0.42}$  & 27.19$_{\pm 0.27}$  \\
\bottomrule
\end{tabular}}
\label{tab:appendix_auc}
\end{table*}

\section{More Experimental Results}\label{sec: more_res}
\begin{figure*}[ht]
    \centering
    \setlength{\belowcaptionskip}{-0.3cm}
    \subfloat[Cora]{\label{fig:roc1}\includegraphics[width=0.25\linewidth]{
    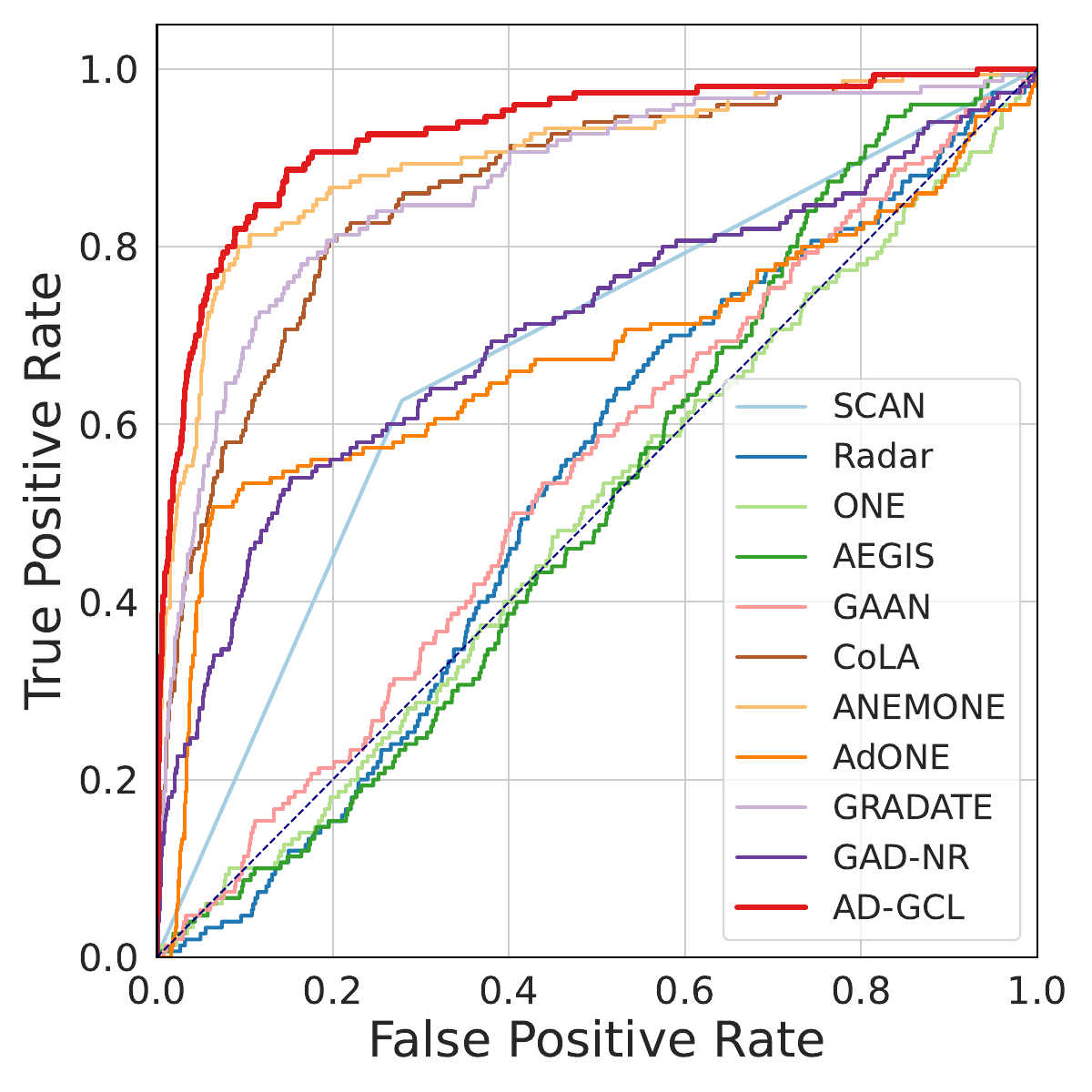}}
    \subfloat[Citeseer]{\label{fig:roc2}\includegraphics[width=0.25\linewidth]{
    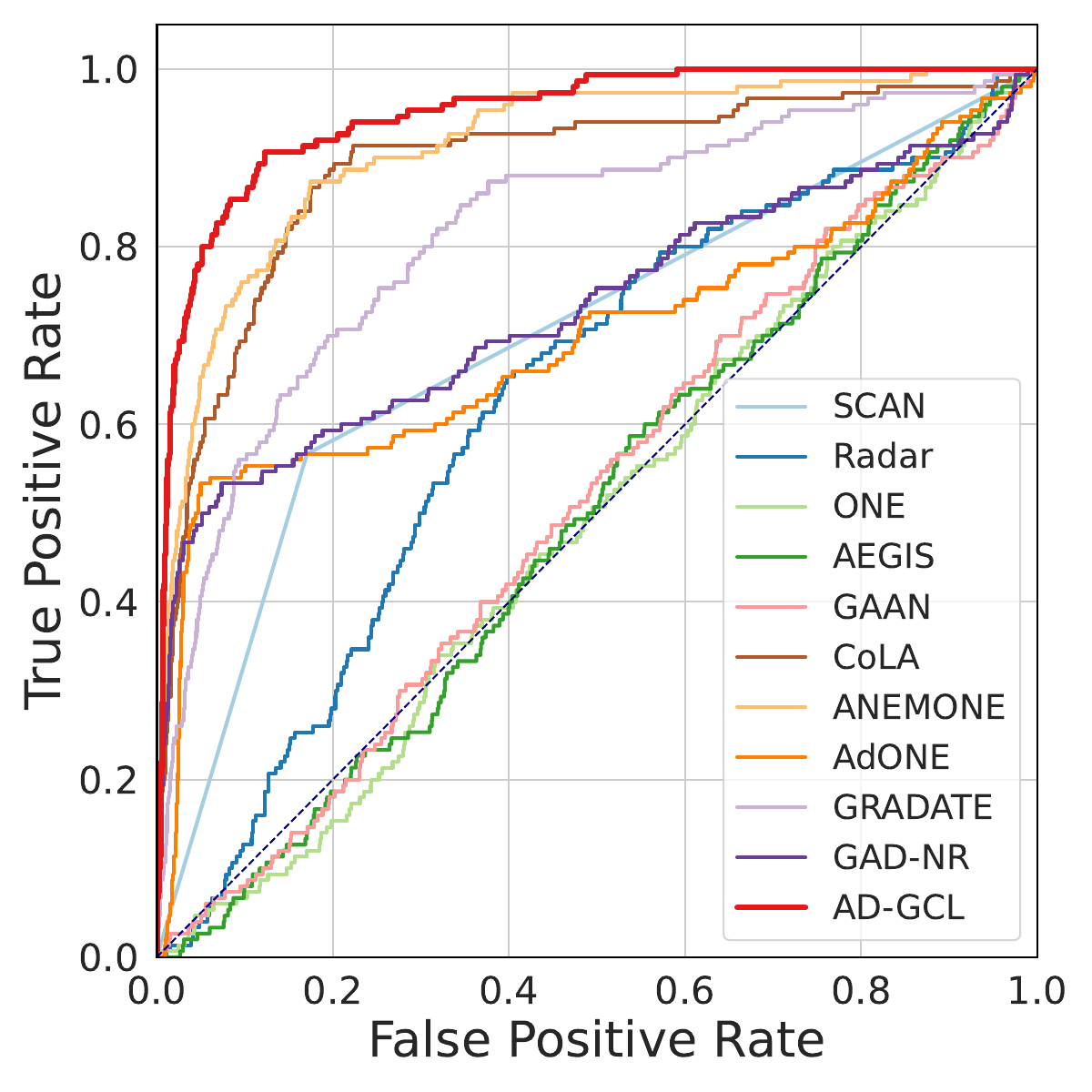}}
    \subfloat[Pubmed]{\label{fig:roc3}\includegraphics[width=0.25\linewidth]{
    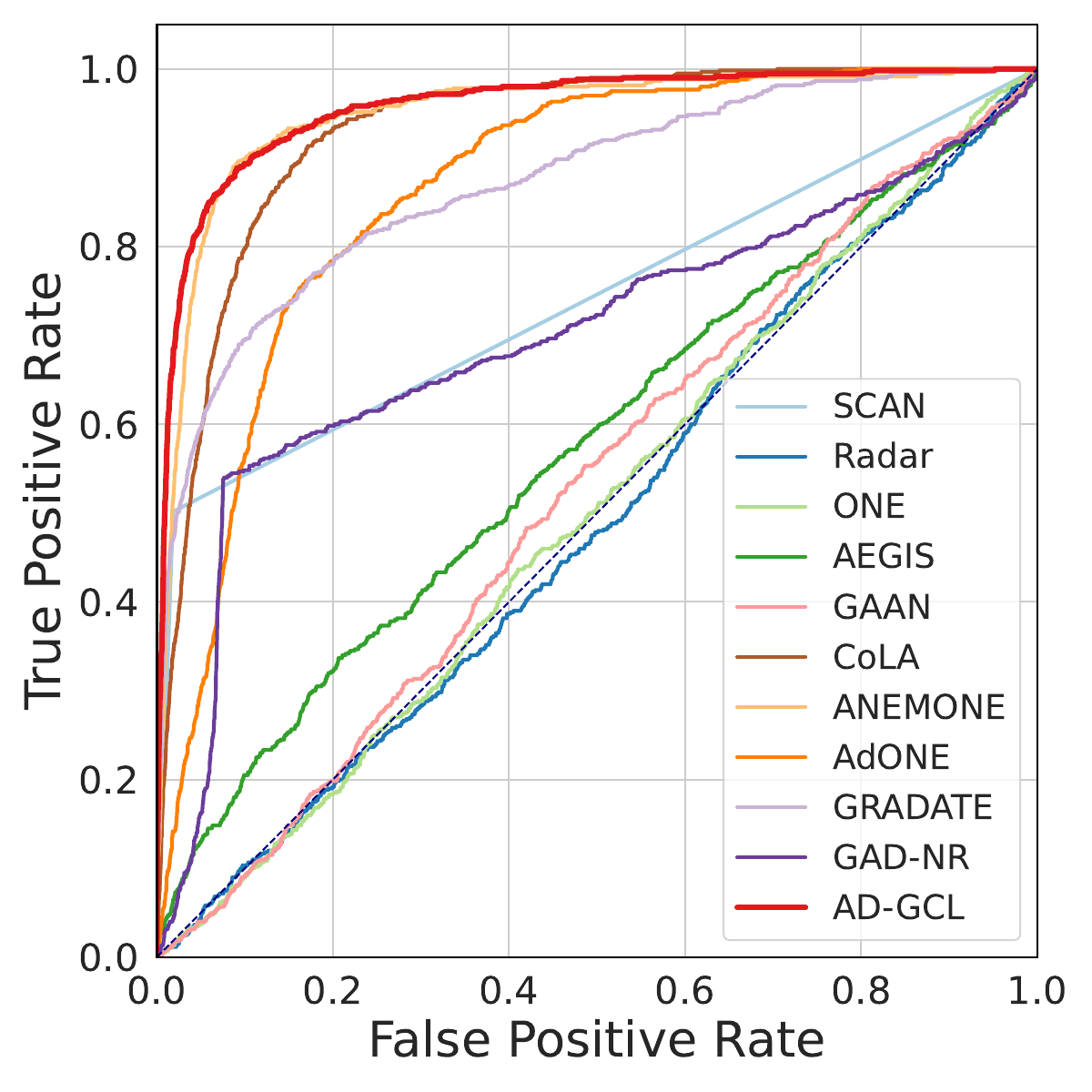}}
    \subfloat[Bitcoinotc]{\label{fig:roc4}\includegraphics[width=0.25\linewidth]{
    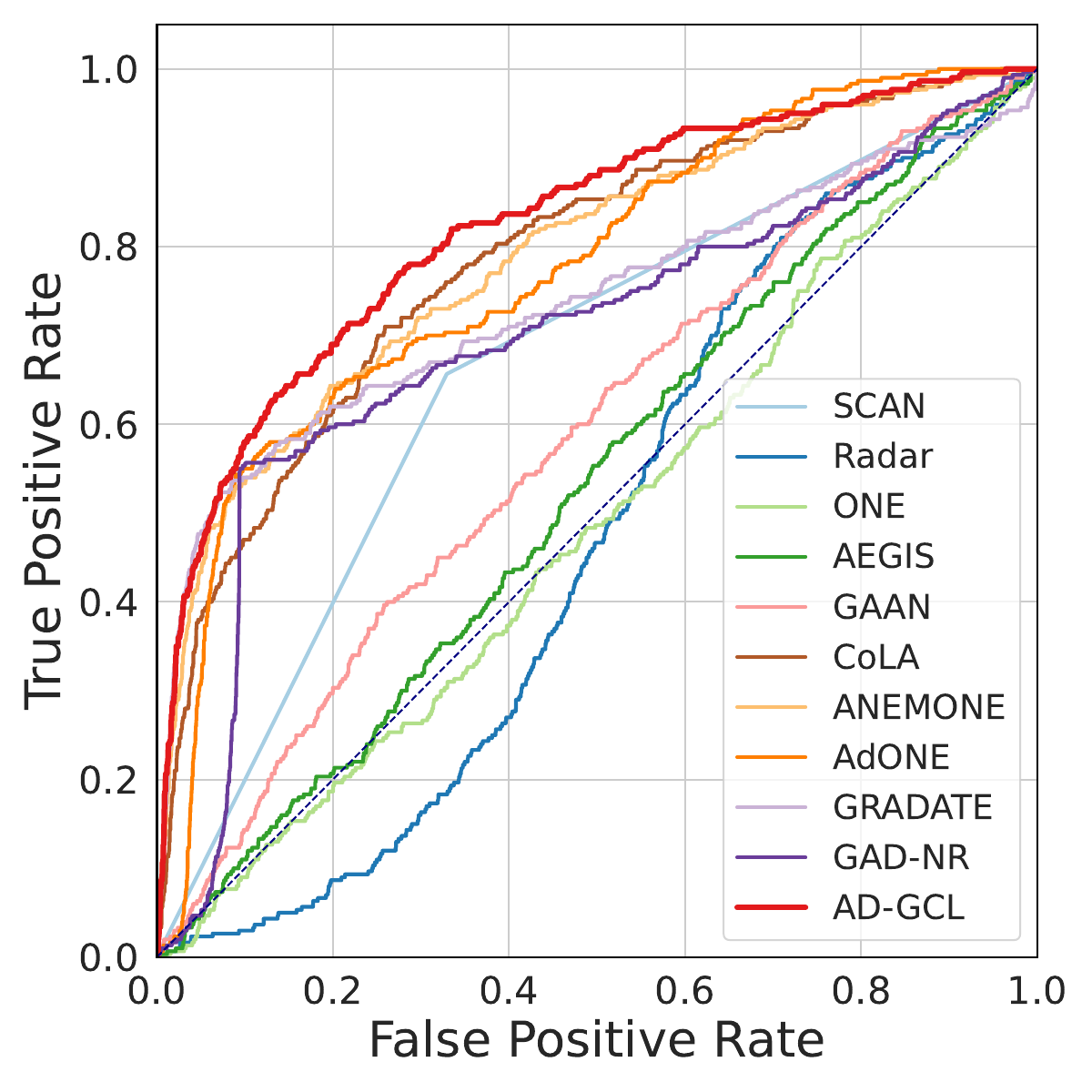}}\\
    \subfloat[BITotc]{\label{fig:roc5}\includegraphics[width=0.25\linewidth]{
    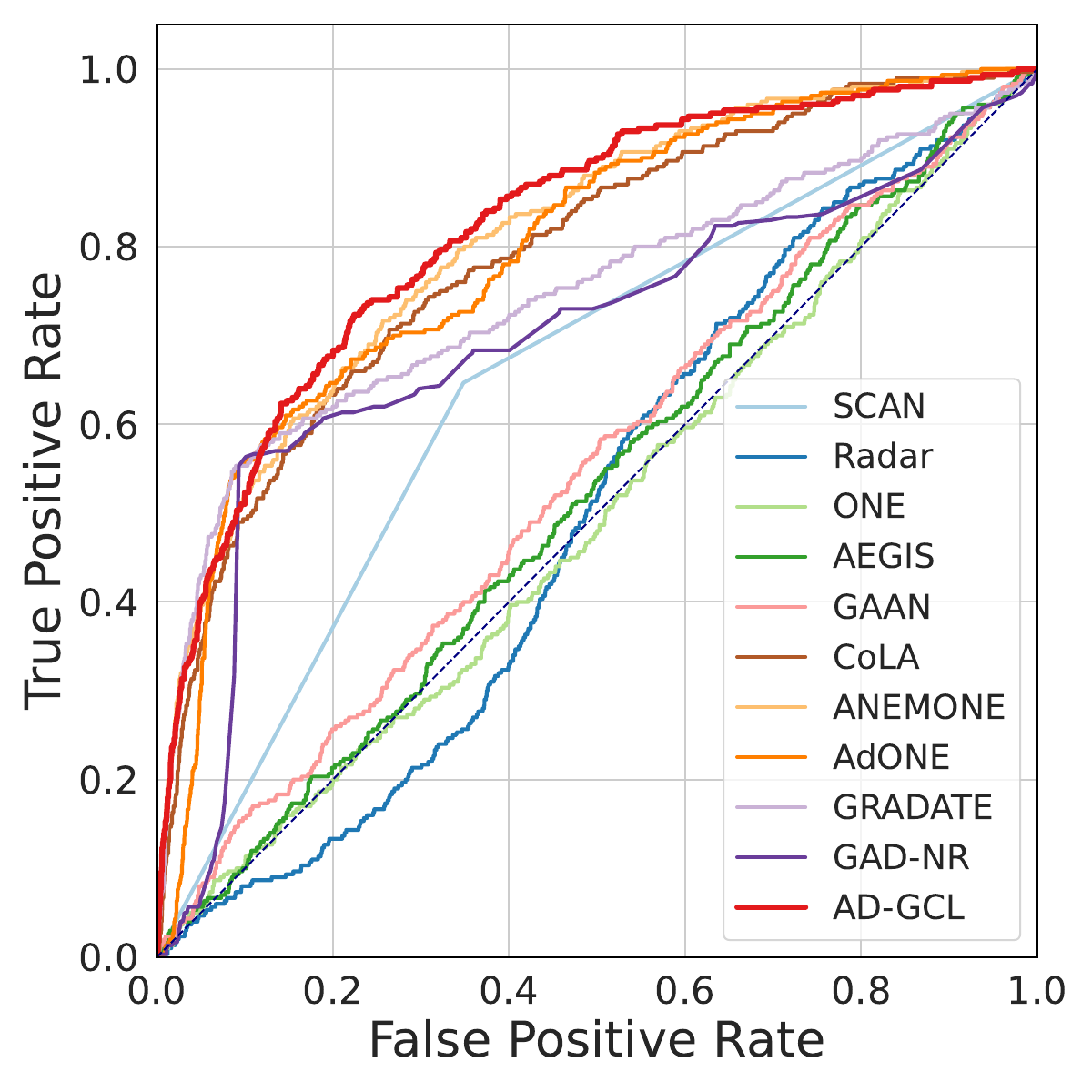}}
    \subfloat[BITalpha]{\label{fig:roc6}\includegraphics[width=0.25\linewidth]{
    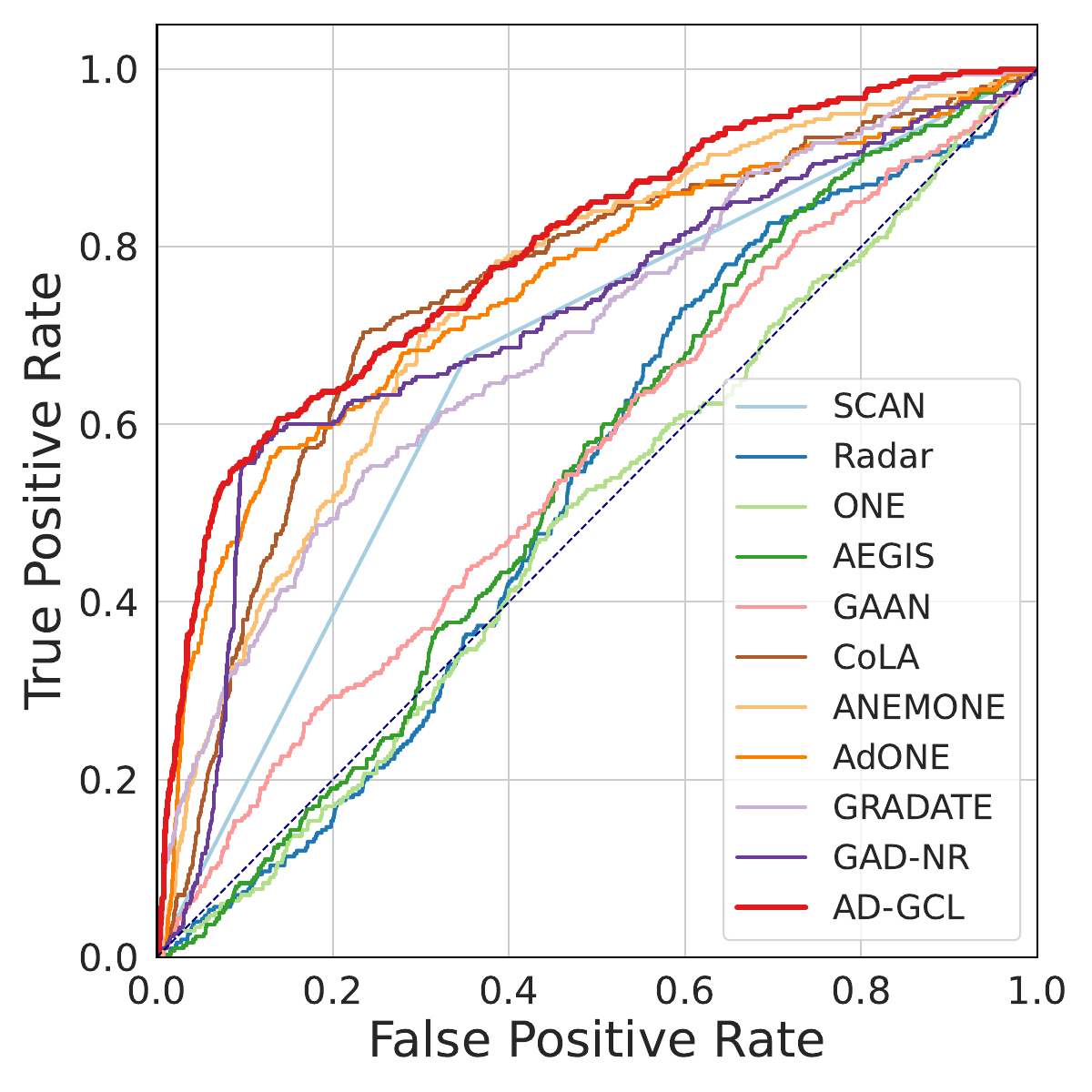}}
    \subfloat[Reddit]{\label{fig:roc7}\includegraphics[width=0.25\linewidth]{
    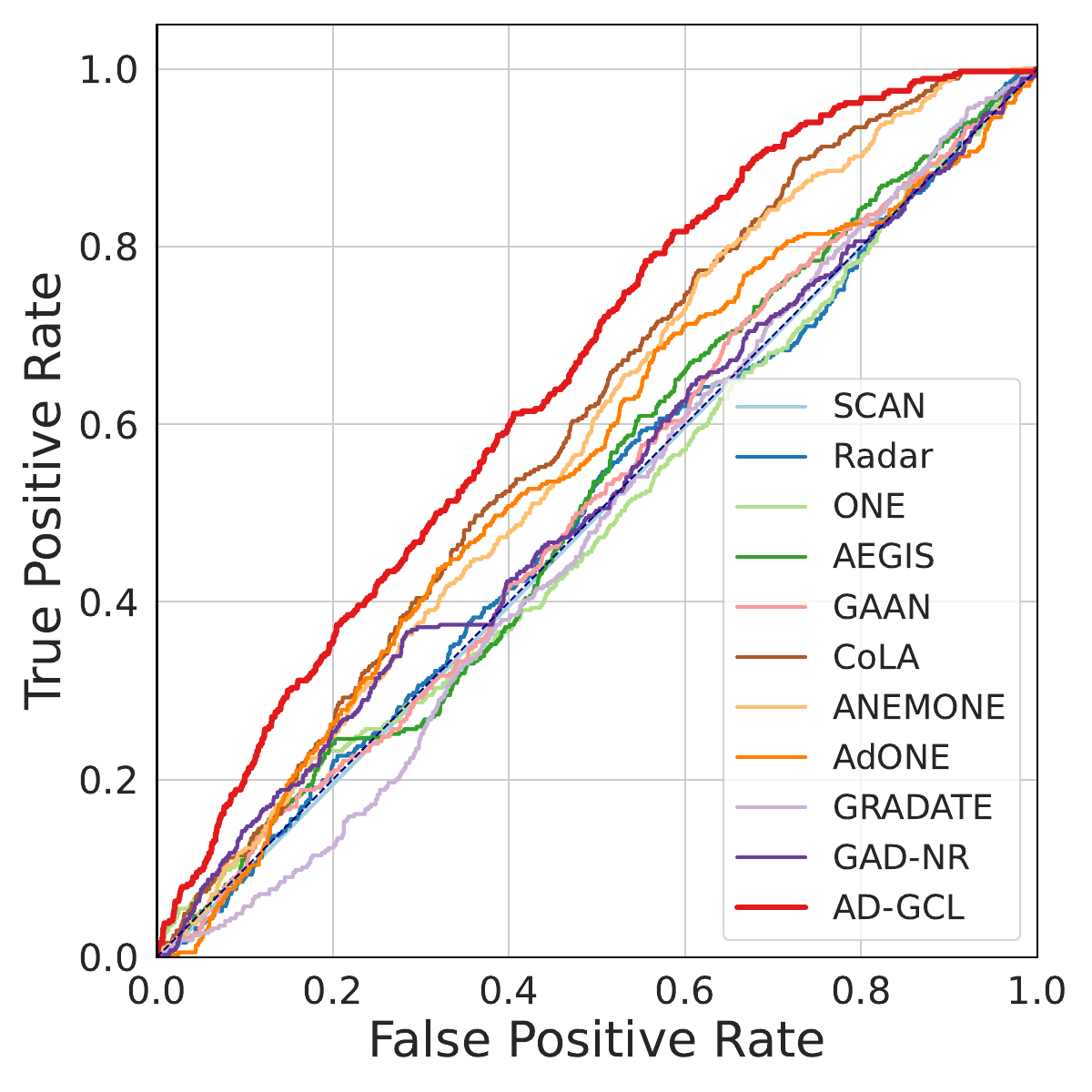}}
    \subfloat[Tolokers]{\label{fig:roc8}\includegraphics[width=0.25\linewidth]{
    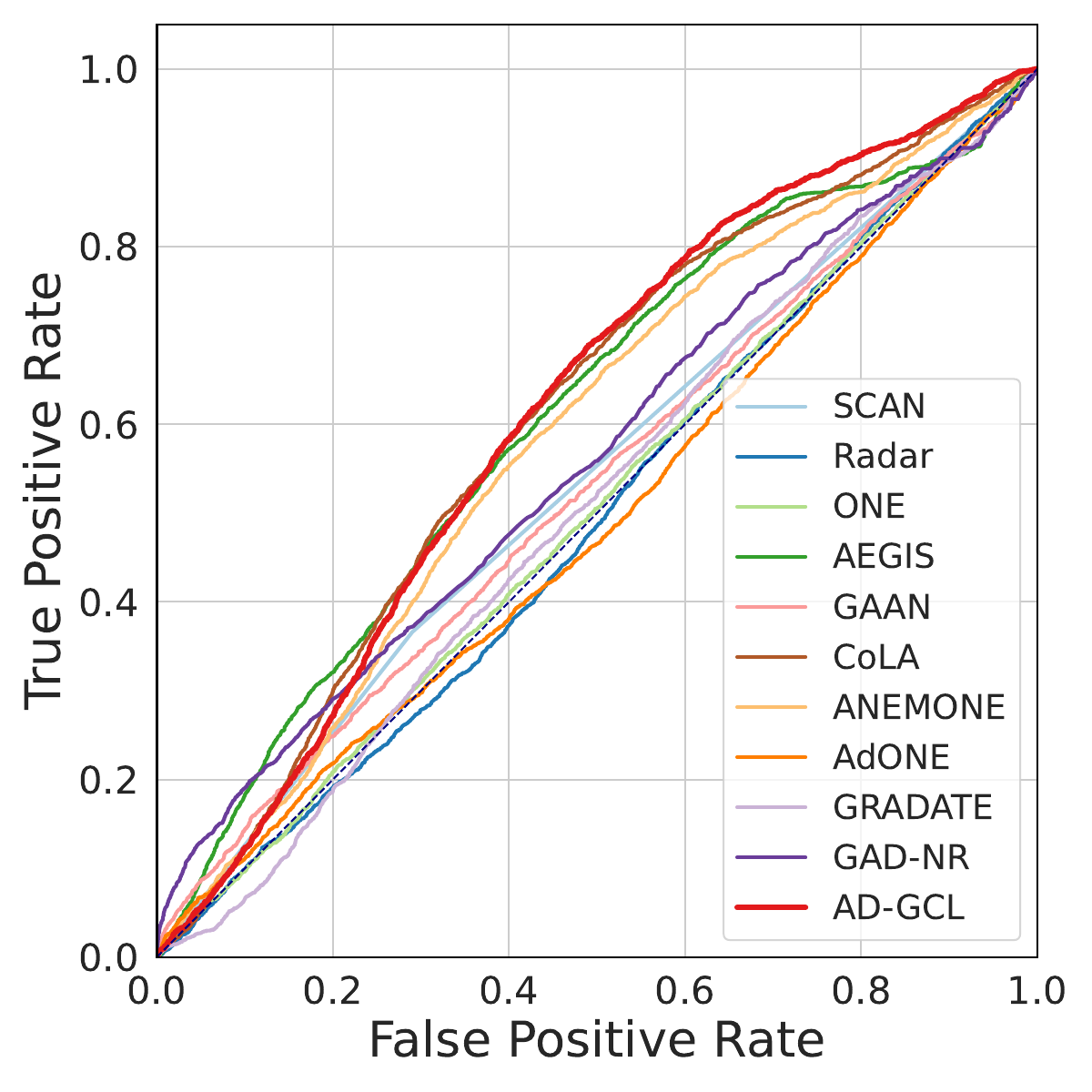}}
    \caption{ROC curves comparison on eight benchmark datasets. A larger area under the curve indicates better performance. The black dotted lines represent the "random line", which indicates the performance achieved through random guessing.}
    \label{fig:roc}    
\end{figure*}

\begin{figure*}[ht]
\centering
\subfloat[CoLA on Cora]{\label{fig:violin1}\includegraphics[width=0.25\linewidth]{
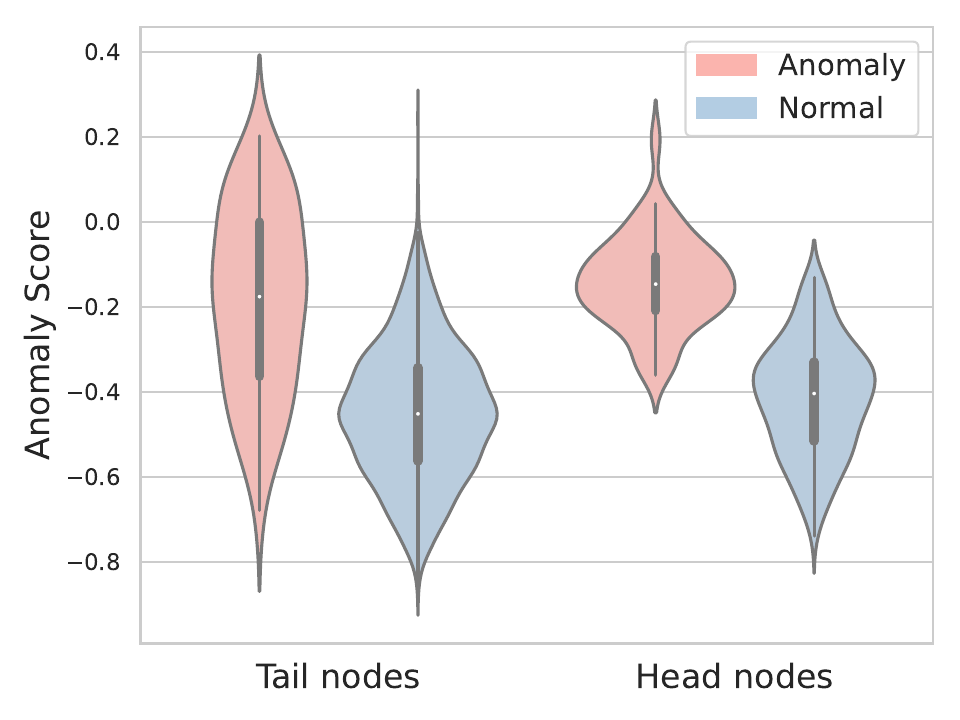}}
\subfloat[ANEMONE on Cora]{\label{fig:violin2}\includegraphics[width=0.25\linewidth]{
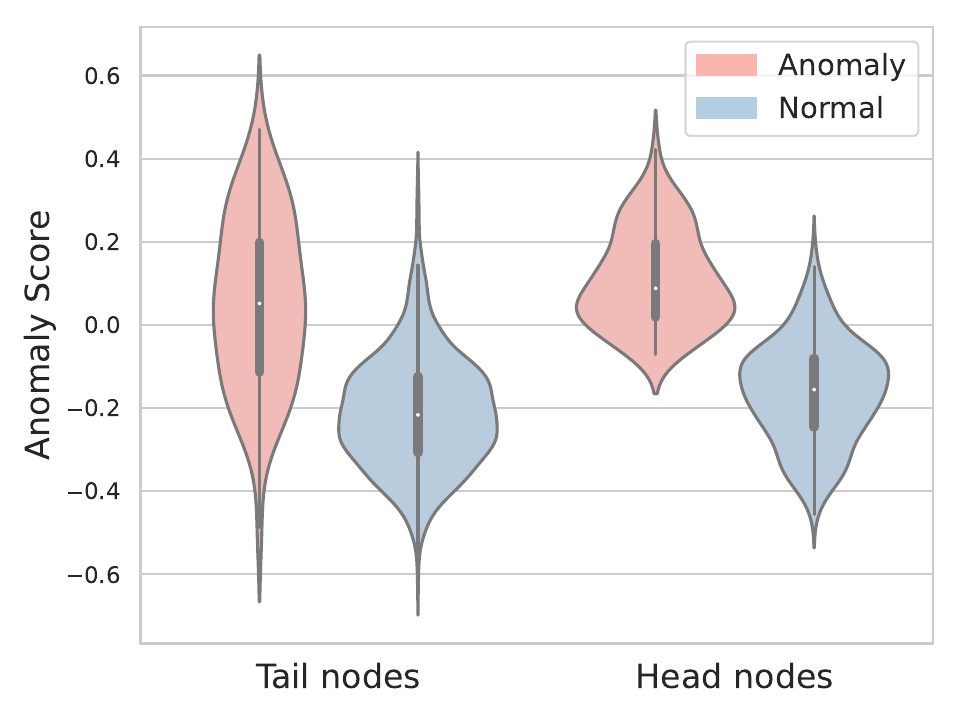}}
\subfloat[GRADATE on Cora]{\label{fig:violin3}\includegraphics[width=0.25\linewidth]{
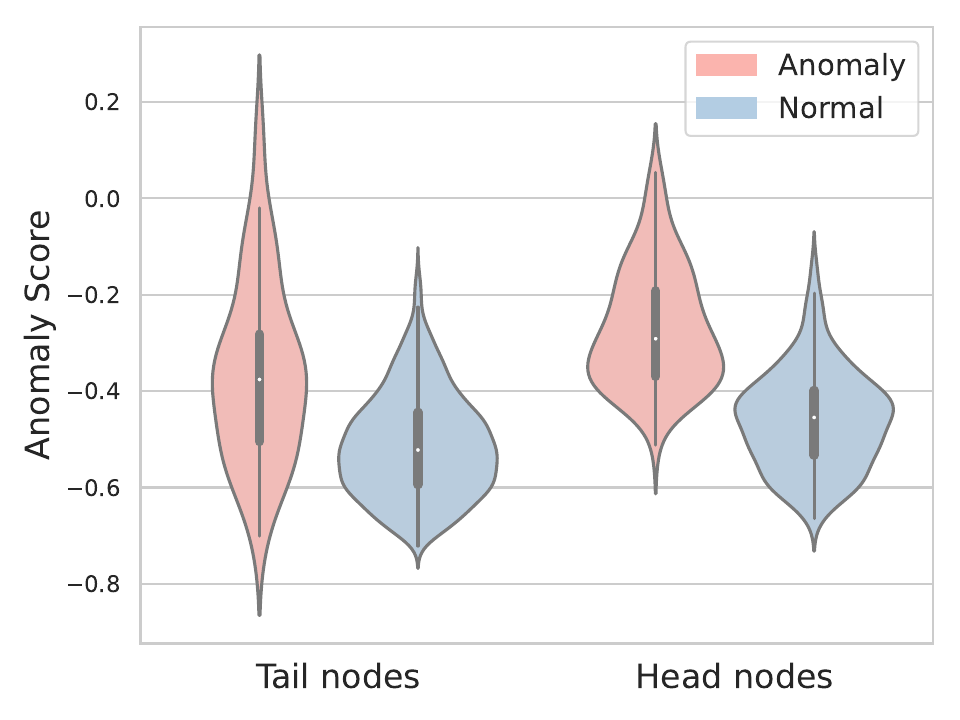}}
\subfloat[AD-GCL on Cora]{\label{fig:violin4}\includegraphics[width=0.25\linewidth]{
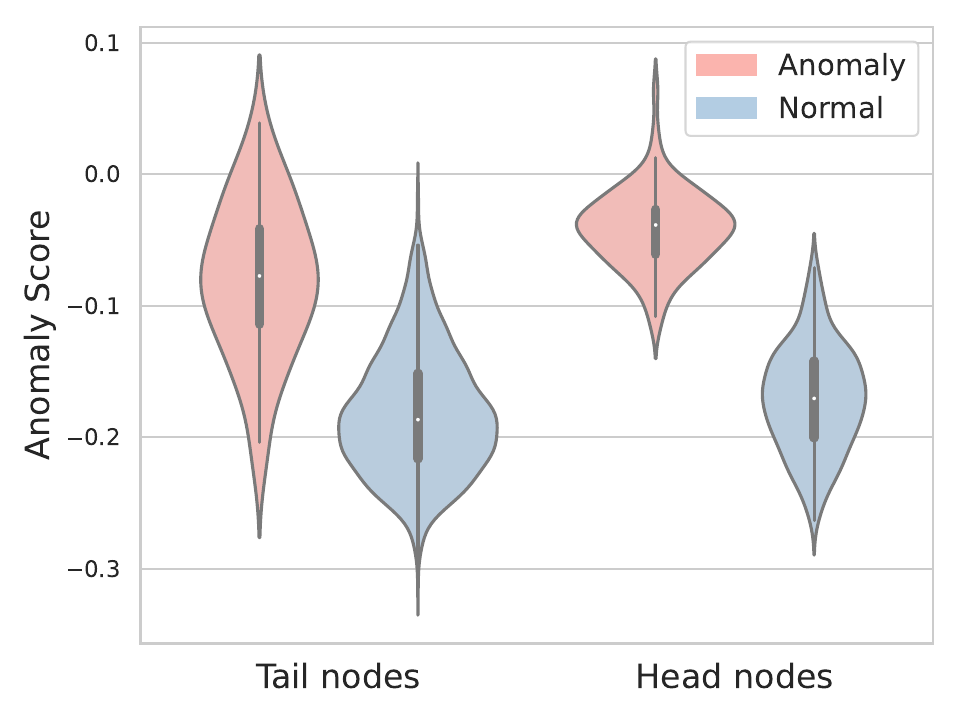}}	\\
\subfloat[CoLA on Citeseer]{\label{fig:violin5}\includegraphics[width=0.25\linewidth]{
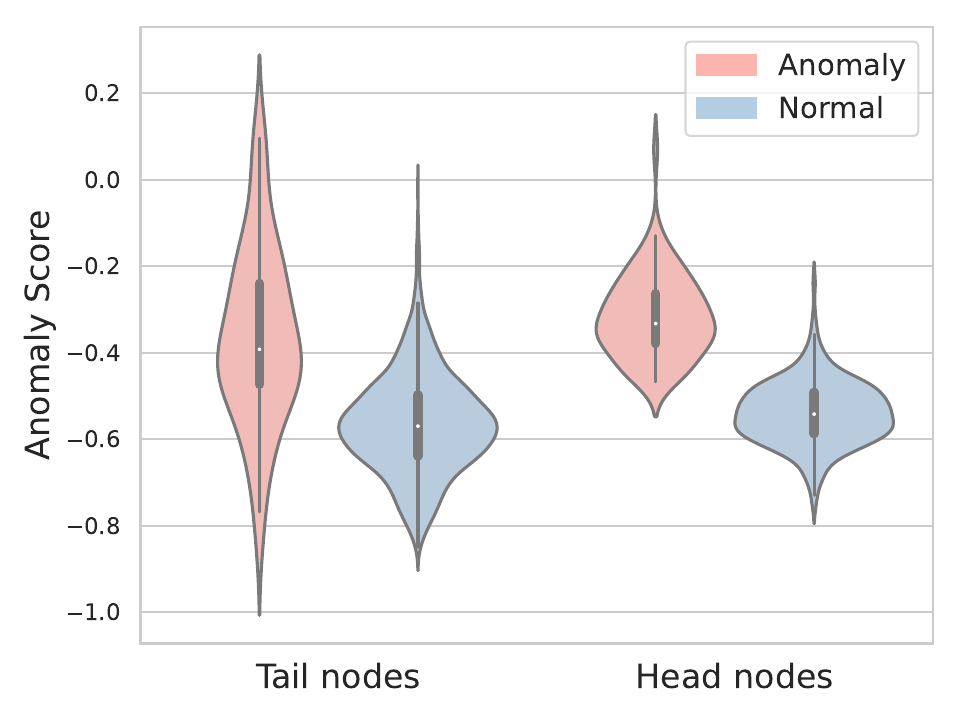}}
\subfloat[ANEMONE on Citeseer]{\label{fig:violin6}\includegraphics[width=0.25\linewidth]{
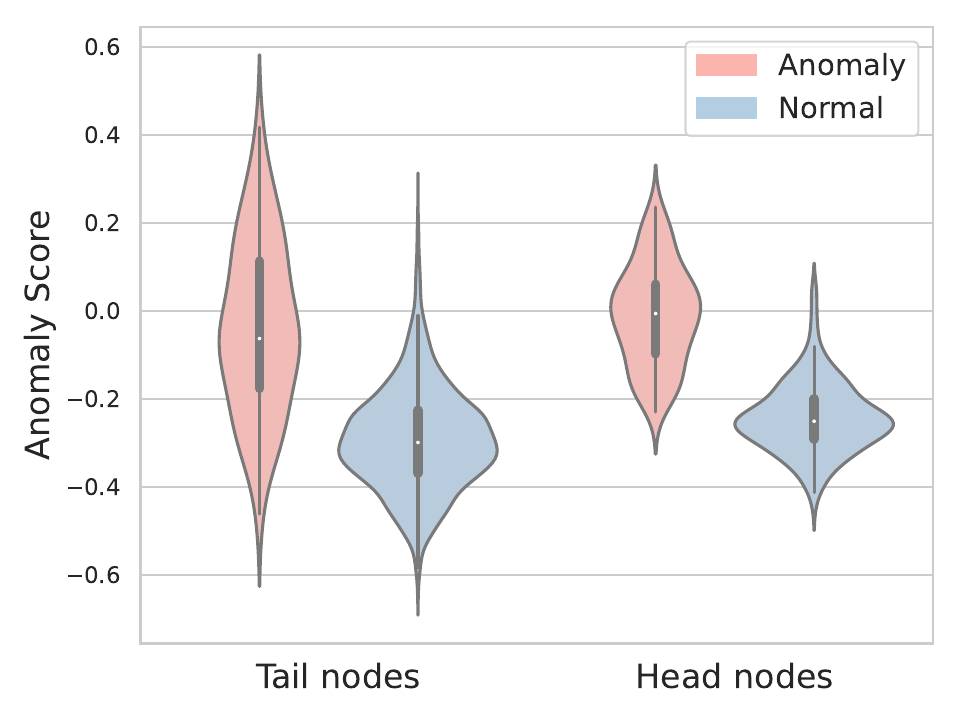}}
\subfloat[GRADATE on Citeseer]{\label{fig:violin7}\includegraphics[width=0.25\linewidth]{
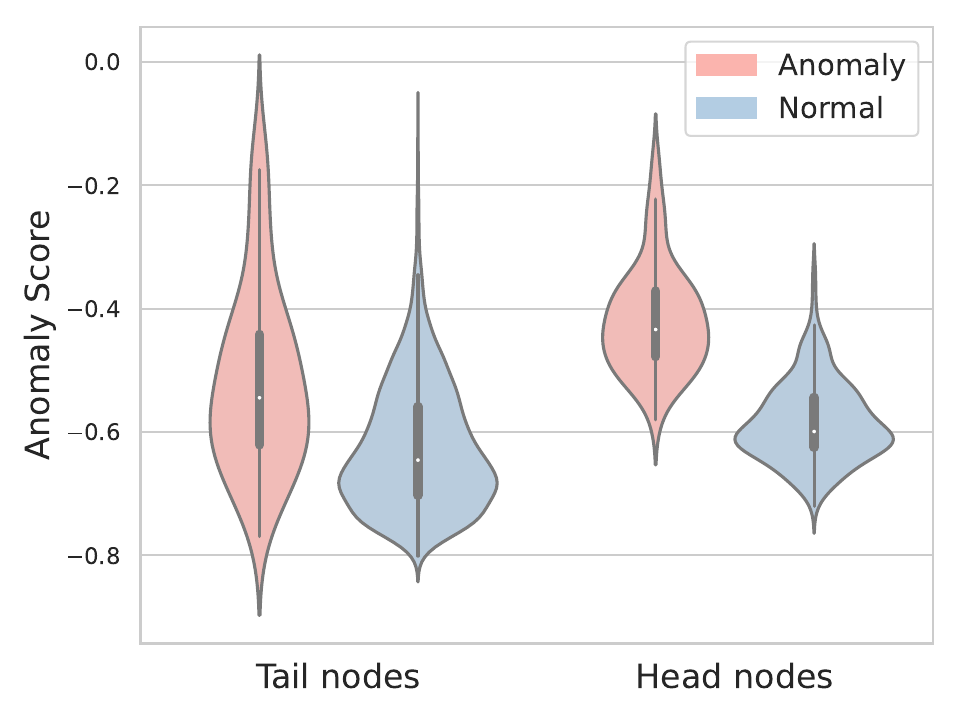}}
\subfloat[AD-GCL on Citeseer]{\label{fig:violin8}\includegraphics[width=0.25\linewidth]{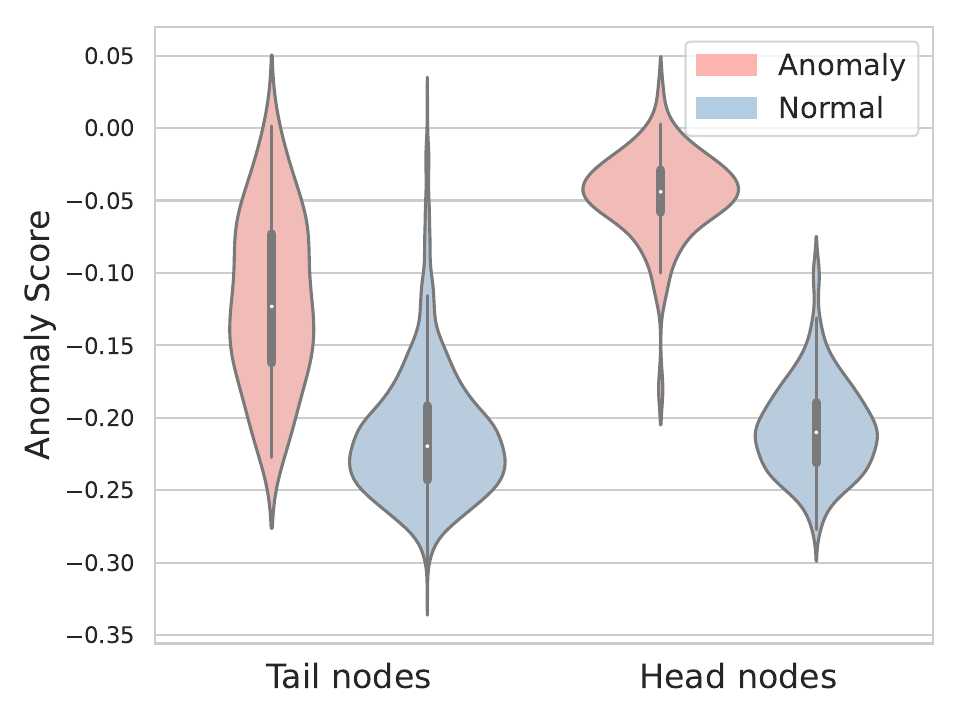}}\\	
\subfloat[CoLA on Reddit]{\label{fig:violin9}\includegraphics[width=0.25\linewidth]{
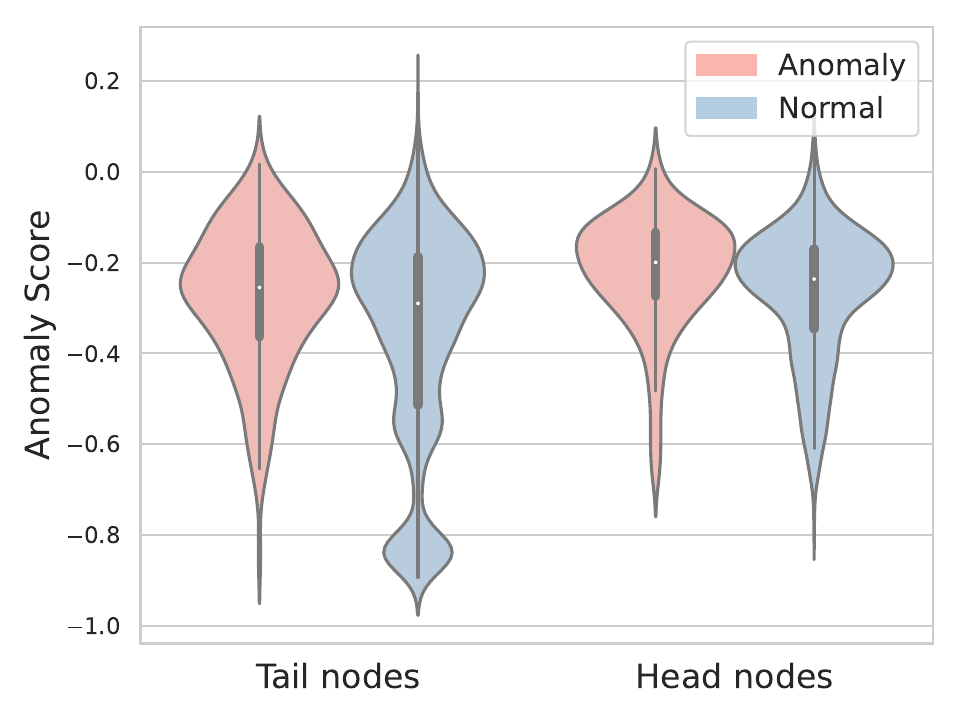}}
\subfloat[ANEMONE on Reddit]{\label{fig:violin10}\includegraphics[width=0.25\linewidth]{
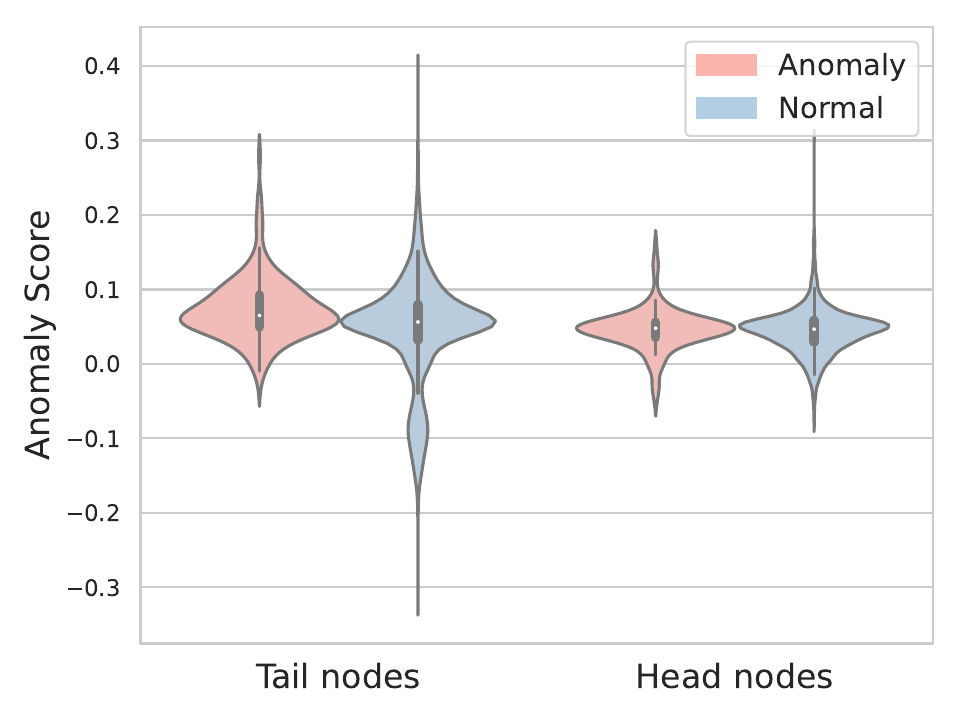}}
\subfloat[GRADATE on Reddit]{\label{fig:violin11}\includegraphics[width=0.25\linewidth]{
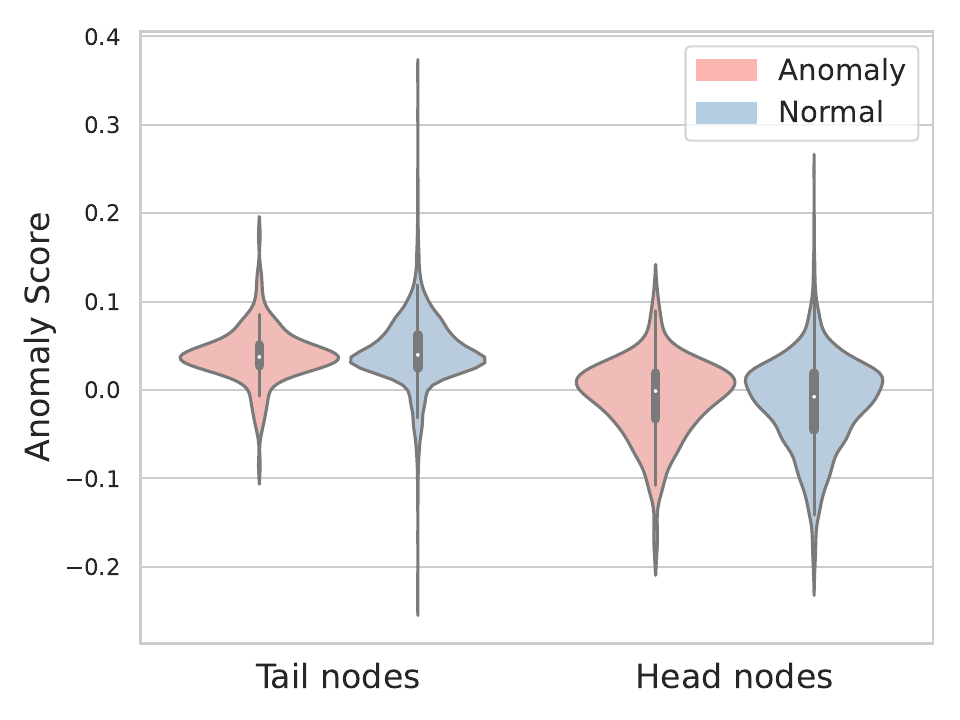}}
\subfloat[AD-GCL on Reddit]{\label{fig:violin12}\includegraphics[width=0.25\linewidth]{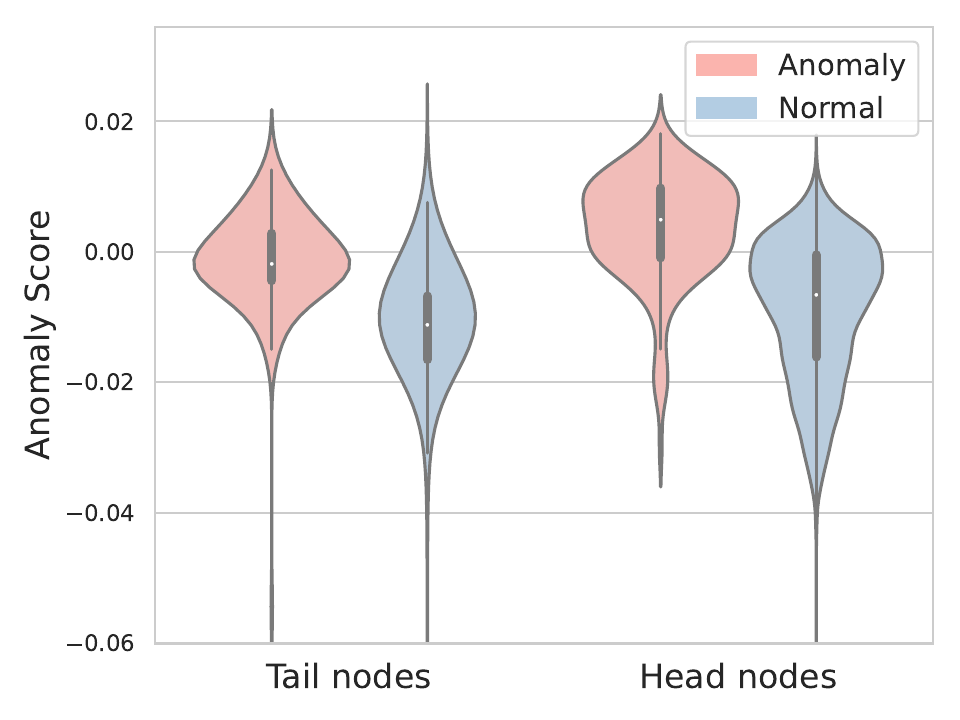}}
\caption{The violin plots illustrate the distribution of model output anomaly scores in both categories of anomaly nodes (red) and normal nodes (blue) on Cora, Citeseer, and Reddit datasets. The width of each violin represents the density of nodes at various anomaly scores, while the vertical lines inside indicate the quartiles and the median is shown by a white scatter.}
\label{fig:violinplot_more}
\end{figure*}

\subsection{More Main Results}\label{sec:more}
We report the experimental results of AUPRC and AP for our proposed method compared to state-of-the-art (SOTA) graph neural network-based approaches in Table~\ref{tab:appendix_auc}.  We can observe that AD-GCL achieves the best results in most cases, whether on all nodes, head nodes, or tail nodes.
Figure~\ref{fig:roc} illustrates the comparison of ROC curves across eight benchmark datasets. A larger area under the curve (AUC) indicates superior performance. Additionally, the black dotted lines represent the "random line," denoting the performance achieved through random guessing. Notably, AD-GCL demonstrates the best performance across all eight datasets. We further supplemented the comparison with three GCL-based anomaly detection methods by providing violin plots of anomaly score distributions on the Cora, Citeseer, and Reddit datasets. As shown in Figure~\ref{fig:violinplot_more}, our method shows more obvious discrimination in the anomaly scores between head anomalies, tail anomalies and normal nodes. This demonstrates that AD-GCL learns more accurate anomaly scores for both normal and anomalous nodes.

\subsection{Compared with Data Augmentation Methods}\label{sec:aug}
To further validate the effectiveness of the neighbor pruning and completion strategies, we replace them with common data augmentation methods, such as edge dropping and feature masking, and random tail node edge completion method. The experimental results show that our method outperforms common data augmentation methods and the naive edge completion method in Table~\ref{tab:aug}.
Traditional graph augmentation often involves random edge dropping and node attribute masking to corrupt the original graph. However, this randomness lacks customized noise filtering for spurious interactions and may discard important edge or feature information, resulting in suboptimal solutions. Similarly, naive tail node completion strategies fail to enhance the performance of tail nodes. Due to the non-i.i.d. nature of graph-structured data, random completion introduces significant noise, resulting in a notable decrease in overall, tail, and head detection performance. In contrast, our proposed neighbor pruning and completion strategies customize the consideration of feature saliency information, neighbor distribution, and anomaly score distribution, making data augmentation more rational.

\begin{table*}[]
\renewcommand\arraystretch{1.3}
\centering
\caption{Performance comparison with common data augmentation methods. TN and HN represent the AUC for tail nodes and head nodes, respectively. Bold indicates the optimal.}
\resizebox{0.81\textwidth}{!}{
\begin{tabular}{ccccccccccc}
\hline
                              & \multirow{2}{*}{Method} & \multicolumn{3}{c}{Cora} & \multicolumn{3}{c}{Citeseer} & \multicolumn{3}{c}{Pubmed} \\
                              \cline{3-11}
                              &                         & AUC     & TN     & HN        & AUC        & TN     & HN    & AUC      & TN      & HN \\ \hline

    & Edge dropping                  
& 91.91 & 83.99 & 98.44 & 91.82 & 84.69 & 98.80 & 95.01 & 94.83 & 97.05\\
    & Feature masking                    
& 91.63 & 83.62 & 97.92 & 92.95 & 86.53 & 98.93 & 94.64 & 93.56 & 97.19\\ 
    & Naive edge completion                    
& 78.28 & 74.85 & 87.87 & 70.33 & 71.37 & 69.45 & 92.03 & 91.69 & 96.47\\ 
    & AD-GCL                                    
& $\mathbf{92.83}$     & $\mathbf{85.70}$        & $\mathbf{98.67}$           & $\mathbf{94.88}$     & $\mathbf{90.51}$       & $\mathbf{99.71}$      & $\mathbf{95.74}$         & $\mathbf{95.12}$       & $\mathbf{97.88}$  \\ 
\hline 
\end{tabular}}
\label{tab:aug}
\end{table*}

\subsection{Multinomial Distribution Sampling}
To further validate the efficacy of multinomial distribution sampling, we substituted it with $k$-nearest neighbor sampling in both the neighbor pruning and neighbor completion phases. The experimental results, depicted in the table~\ref{tab:mds}, illustrate that multinomial distribution sampling outperformed $k$-nearest neighbor sampling across all six datasets, underscoring the effectiveness of multinomial distribution sampling. $K$-nearest neighbor sampling typically selects a fixed number ($k$) of nearest neighbors, which can be sensitive to noisy data. In contrast, our proposed multinomial distribution sampling assigns sampling probabilities based on node importance, resulting in more robust and generalized sampling outcomes. Just as dropout regularization introduces randomness during training to prevent overfitting by encouraging the network to learn more robust features, our multinomial distribution sampling approach similarly introduces diversity in the sampling process.

\begin{table*}[t!]
\renewcommand\arraystretch{1.2}
\centering
  \caption{Comparison of multinomial distribution sampling and k-nearest neighbor sampling in neighbor pruning and completion.}
  \label{tab:mds}
  \resizebox{0.81\textwidth}{!}{
  \begin{tabular}{ccccccc}
    \toprule
    Readout function & Cora & Citeseer & Pubmed & Bitcoinotc & BITotc & BITalpha \\
    \midrule
    $K$-nearest neighbor sampling & 90.33 & 93.15 & 95.36 & 80.38 & 80.85 & 79.08 \\
    Multinomial distribution sampling (default) & $\mathbf{92.83}$ & $\mathbf{94.88}$ & $\mathbf{95.74}$ & $\mathbf{82.19}$ & $\mathbf{82.11}$ & $\mathbf{79.62}$ \\
    \bottomrule
  \end{tabular}}
\end{table*}

\begin{table*}
\renewcommand\arraystretch{1.2}
\centering
  \caption{Sensitivity of the Readout Function.}
  \label{tab:readout}
  \begin{tabular}{ccccccc}
    \toprule
    Readout function & Cora & Citeseer & Pubmed & Bitcoinotc & BITotc & BITalpha \\
    \midrule
    MAX & 91.54 & 94.08 & 94.71 & 81.81 & 81.28 & 78.11 \\
    MIN & 92.57 & 93.77 & 95.36 & 81.28 & 79.76 & 78.34 \\
    MEAN (default) & 92.83 & 94.88 & 95.74 & 82.19 & 82.11 & 79.62 \\
    \bottomrule
  \end{tabular}
\end{table*}

\subsection{Sensitivity of the Readout Function}
We evaluate the sensitivity of AD-GCL to different readout functions in Eq.~\eqref{eq:readout}, including the common MEAN (default), MAX, and MIN functions. The experimental results, as shown in Table~\ref{tab:readout}, indicate that all three readout functions perform competitively well. AD-GCL demonstrates strong robustness across different readout functions.

\section{Limitations}\label{sec:limitations}
Although we have demonstrated the effectiveness of AD-GCL on multiple real-world datasets, we also notice some limitations due to the complexity of graph anomaly detection and learning from structural imbalances. First, our method focuses on homophilic graphs (linked nodes should be similar) that are ubiquitous in real-world scenarios. However, the effectiveness of our method in heterophilic graphs (linked nodes are expected to be distinct) needs further confirmation. Second, the current degree threshold is defined based on previous work and the Pareto principle. It might be beneficial to explore a self-adaptive mechanism to dynamically adjust the threshold for each graph. Thirdly, using theoretical insights to improve the lower and upper bounds of the model and guide its design will be an interesting direction for future research. 

\end{document}